\documentclass{article}

\usepackage[utf8]{inputenc}
\usepackage[english]{babel}
\usepackage[T1]{fontenc}

\usepackage{amsthm}
\usepackage{mathrsfs}
\usepackage{fancyhdr}
\usepackage[colorlinks,linkcolor=red,citecolor=blue,urlcolor=black,filecolor=blue,backref=page]{hyperref}
\usepackage{graphicx}
\usepackage{amsmath}
\usepackage{amssymb}
\usepackage{dsfont}

\usepackage{appendix}
\usepackage{etoc}

\usepackage{float} 
\usepackage{natbib}

\usepackage{authblk}
\usepackage{fullpage}

\theoremstyle{plain}
\newtheorem{theorem}{Theorem}[section]

\theoremstyle{remark}

\newtheorem{example}[theorem]{\bf{Example}}

\newcommand{\prob}{\mathbb{P}}
\newcommand{\mean}{\mathbb{E}}

\newcommand{\var}{\text{var}}

\newcommand{\ud}{\mathop{}\!\mathrm{d}}
\newcommand{\cond}{\,|\,}

\DeclareMathOperator{\Normal}{\mathcal{N}}

\DeclareMathOperator{\Poi}{{Poi}}
\DeclareMathOperator{\GM}{\mathcal{GM}}

\DeclareMathOperator{\GP}{\mathcal{GP}}
\DeclareMathOperator{\Unif}{\mathcal{U}}

\newcommand{\Bk}{\mathbf{k}}

\newcommand{\Bx}{\mathbf{x}}
\newcommand{\By}{\mathbf{y}}

\newcommand{\BK}{\mathbf{K}}

\newcommand{\BSigma}{\boldsymbol{\Sigma}}

\newcommand{\Btheta}{\boldsymbol{\theta}}
\newcommand{\BTheta}{\boldsymbol{\Theta}}

\newcommand{\Bphi}{\boldsymbol{\phi}}

\renewcommand{\epsilon}{\varepsilon}

\newcommand{\reals}{\mathbb{R}}

\newcommand{\mlpd}{mlpd utility}
\newcommand{\cmlpd}{classifier utility}

\newcommand{\classgp}{classifier GP}

\newcommand{\abcrej}{rej. ABC}
\newcommand{\discr}{\Delta}

\usepackage{booktabs} 
\newcommand{\ra}[1]{\renewcommand{\arraystretch}{#1}} 

\def\app#1#2{%
  \mathrel{%
    \setbox0=\hbox{$#1\sim$}%
    \setbox2=\hbox{%
      \rlap{\hbox{$#1\propto$}}%
      \lower1.1\ht0\box0%
    }%
    \raise0.25\ht2\box2%
  }%
}
\def\approxprop{\mathpalette\app\relax}

\newcommand{\indic}{\mathds{1}}

\allowdisplaybreaks 

\begin{document}

\title{Gaussian process modeling in approximate Bayesian computation to estimate horizontal gene transfer in bacteria}

\author[1]{Marko Järvenpää}
\author[2]{Michael U. Gutmann}
\author[1]{Aki Vehtari}
\author[1]{Pekka Marttinen}

\affil[1]{Helsinki Institute for Information Technology HIIT, Department of Computer Science, Aalto University}
\affil[2]{School of Informatics, University of Edinburgh}

\date{\today}

\maketitle

\begin{abstract}
Approximate Bayesian computation (ABC) can be used for model fitting when the likelihood function is intractable but simulating from the model is feasible. However, even a single evaluation of a complex model may take several hours, limiting the number of model evaluations available. Modelling the discrepancy between the simulated and observed data using a Gaussian process (GP) can be used to reduce the number of model evaluations required by ABC, but the sensitivity of this approach to a specific GP formulation has not yet been thoroughly investigated. 
We begin with a comprehensive empirical evaluation of using GPs in ABC, including various transformations of the discrepancies and two novel GP formulations. Our results indicate the choice of GP may significantly affect the accuracy of the estimated posterior distribution. Selection of an appropriate GP model is thus important. We formulate expected utility to measure the accuracy of classifying discrepancies below or above the ABC threshold, and show that it can be used to automate the GP model selection step. Finally, based on the understanding gained with toy examples, we fit a population genetic model for bacteria, providing insight into horizontal gene transfer events within the population and from external origins. 
\end{abstract}


\section{Introduction} \label{sec:intro}

Estimating parameters of a statistical model often requires evaluating the likelihood function. For complex models, such as those arising in population genetics, deriving or evaluating the likelihood in a reasonable computation time may be impossible. On the other hand, generating data from the model may be relatively straightforward. Approximate Bayesian Computation (ABC) \citep{Beaumont2002,Hartig2011,Marin2012,Turner2012,Lintusaari2016} is an inference framework for such models. It is based on generating data from the simulation model for various parameter values and comparing the simulated data with the observed data using some discrepancy measure. The simplest ABC algorithm is the rejection sampler, which, at each step, randomly simulates a parameter from the prior distribution, runs the simulation model with this parameter, and finally accepts the parameter if the discrepancy between the simulated and observed data is smaller than some threshold parameter (which we call ``ABC threshold'' or just ``threshold''). These steps are repeated until a sufficient number of samples from the approximate posterior have been collected.

To speed up ABC inference, several sampling-based algorithms have been proposed \citep{Marjoram2003,Sisson2007,Beaumont2009,Toni2009,Drovandi2011,Moral2012,Lenormand2013}. 
An alternative to sampling that has received much attention in recent years is to construct an explicit approximation to the likelihood function, and use this as a proxy for the exact likelihood in e.g. MCMC samplers. 
In the synthetic likelihood method this is done by modelling the summary statistics with a multivariate Gaussian \citep{Wood2010,Price2016}, see also \citet{Fan2013,Papamakarios2016} for some other approaches. Nonparametric approximations have also been considered \citep{Blum2010,Turner2014}, and connections to other approaches are discussed by \citet{Drovandi2015,Gutmann2015}. Gaussian processes \citep{Rasmussen2006} (GPs) can naturally encode assumptions about the smoothness of the likelihood. They have been used by \citet{Drovandi2015b} to accelerate pseudo-marginal MCMC methods, and by \citet{Wilkinson2014,Kandasamy2015} to model the likelihood function. An alternative is to model individual summaries with a GP \citep{Meeds2014,Jabot2014}.

Typically hundreds of thousands of model simulations are needed for ABC inference, but here we focus on the challenging case where less than a thousand evaluations are available due to computational constraints. We adopt the approach of \citet{Gutmann2015} who modelled the discrepancy between observed and simulated data with a GP. 
In this paper, by discrepancy we mean a scalar-valued non-negative function that measures the distance between the observed and simulated data.
Modelling the scalar-valued discrepancy allows one to use Bayesian optimisation \citep{Brochu2010,Shahriari2015} to effectively select evaluation locations \citep{Gutmann2015}. Also, this approach has the advantage that computing the ABC posterior estimate can be done even with relatively few model evaluations. 
The ABC posterior is proportional to the product of the prior and the probability that the simulated discrepancy falls below the ABC threshold, and this quantity can be computed analytically from the fitted GP. However, a potential issue in using a GP to model the discrepancy is that, in practice, the GP modelling assumptions may not hold exactly \citep{Gutmann2015}. The discrepancy is often positive (e.g. a weighted Euclidean distance), non-Gaussian, and its variance may vary over the parameter space, causing additional approximation error of unknown magnitude. In this article we study this in detail. To focus on the GP modelling aspect, we assume that the region of non-negligible posterior probability is known in advance, but acknowledge that detecting the region is a topic of ongoing research on its own \citep{Wilkinson2014,Kandasamy2015,Drovandi2015b,Gutmann2015,Jarvenpaa2017}.

\begin{figure}[ht]
\centering
\includegraphics[width=0.65\textwidth]{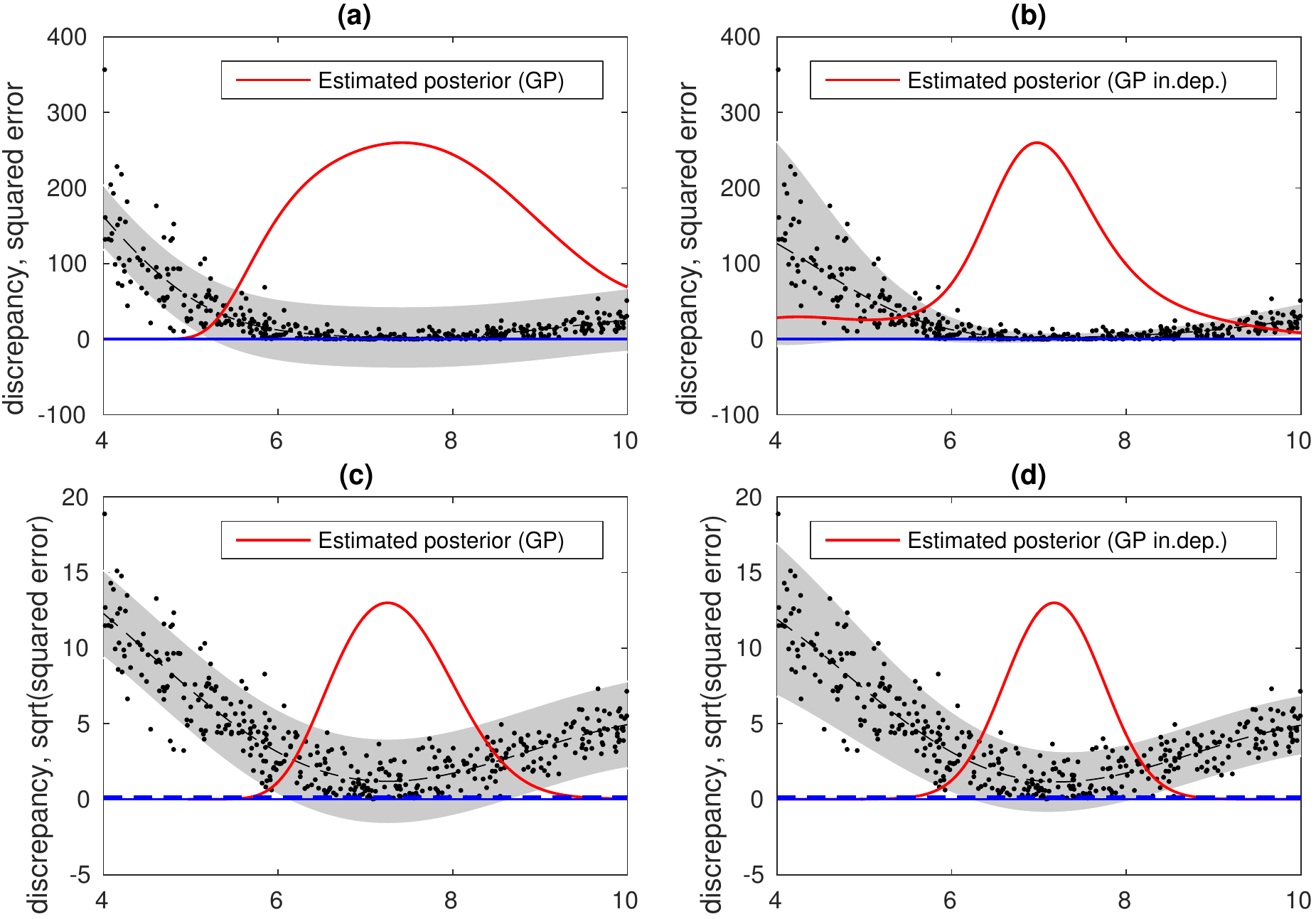}
\caption
{The GP surrogate used to model the simulated discrepancies affects the accuracy of the resulting ABC posterior estimate. 
$x$-axis is the simulation model parameter and $y$-axis the value of the (transformed) discrepancy. Black dots are the simulated realisations of discrepancies, and the grey area is the 95\% predictive interval, representing stochastic variation in the simulation. The red line shows the corresponding posterior approximation which is computed as a lower tail probability from the discrepancy model. (a) The standard GP results in overestimated variance of the discrepancy, yielding a poor approximation to the posterior. Input-dependent GP model in (b) or discrepancy transformation in (c) result in better approximations. The best fit is here obtained when using both the transformation and the input-dependent GP model in (d), as even after the square-root transformation in (c) the variance of the discrepancy is not constant.} \label{fig:intro_demo}
\end{figure}

The impact of GP model assumptions on the resulting ABC posterior is demonstrated with a realistic example in Figure \ref{fig:intro_demo}, where different GP formulations are used to model the discrepancy in the area with non-negligible posterior probability. The model here describes horizontal gene transfer between bacterial genomes, published recently by \citet{Marttinen2015}. The discrepancies were obtained by fixing other parameters to their point estimates, and generating realisations of the discrepancy with varying values for a parameter that describes the frequency of horizontal gene transfer between bacteria. A thorough analysis of the model is presented in Section \ref{subsec:real_examples}. For now please note that the input-dependent noise model \citep{Goldberg1997,Tolvanen2014} is able to take into account the heteroscedastic variance of the discrepancy and, consequently, seems to result in an accurate approximation to the posterior (the true posterior is here unavailable). On the other hand, with the standard GP regression the fit is poor, and the resulting posterior distribution appears too wide.

Our paper makes the following contributions: 
\begin{itemize}
    \item Motivated by the preliminary investigation with the population genetics model above, we assess the impact of the GP formulation on ABC inference for multiple benchmark models. 
    \item We propose two generalisations of previously presented GP-ABC approaches: first, we allow heteroscedastic noise in the GP; second, we use a classifier GP to directly model the probability of the discrepancy being below the ABC threshold.
    \item We propose a new utility function to automate GP model choice for ABC. The utility function favours models that achieve higher accuracy in classifying discrepancies below or above the ABC threshold.
    \item As a practical application, we derive an accurate posterior distribution for the population genetic model for gene transfer in bacteria, allowing us to make inferences about the relationship between gene deletions and introductions, and between gene transfers from within the population and from external origins.
\end{itemize}

This paper is organised as follows. In Section \ref{sec:lfi} we briefly review general ABC methods and introduce different GP models for ABC. We also discuss GP model selection in ABC. 
In Section \ref{sec:examples} we present findings from multiple example problems to illustrate the impact of GP assumptions and model selection in ABC, and finally present the results for the bacterial genomics model. Section \ref{sec:discussion} contains discussion and in Section \ref{sec:concl} we conclude with recommendations on handling GP surrogates in ABC inference.

\section{Background and methods} \label{sec:lfi}

\subsection{ABC} \label{subsec:abc}

We assume that we have observed data $\By\in\reals^d$ from a simulation model whose likelihood function can be written as $p(\By\cond\Btheta)$, where the unknown parameters to be estimated are $\Btheta \in \Theta \subset \reals^p$, and the prior density is $p(\Btheta)$. The posterior distribution can then be computed from the Bayes' theorem 
\begin{align}
p(\Btheta \cond \By) = \frac{p(\Btheta)p(\By\cond\Btheta)}{\int p(\Btheta')p(\By\cond\Btheta') \ud \Btheta'} \propto p(\Btheta)p(\By\cond\Btheta).
\end{align}
When either the analytic form of the likelihood function $p(\By\cond\Btheta)$ is unavailable or its value cannot be evaluated in a reasonable time, the standard alternative is to use approximate Bayesian computation (ABC). 
The ABC targets the approximate posterior  
\begin{align}
p_{\text{ABC}}(\Btheta \cond \By) \propto p(\Btheta)\int \indic_{\discr(\By,\Bx) \leq \epsilon} \, p(\Bx\cond\Btheta) \ud \Bx,
\end{align}
where $\Bx\in\reals^d$ denotes pseudo-data generated by the simulation model with parameter $\Btheta$. The pseudo-data $\Bx$ are compared to the observed data $\By$ 
and $\discr: \reals^d \times \reals^d \rightarrow \reals_{+}$ is a discrepancy function between the two data sets. In practice, the threshold $\epsilon$ represents a tradeoff between estimation accuracy and efficiency; small values result in more accurate estimates but require more computation. The discrepancy is often formed using some summary statistics such that if $s$ is a mapping from the data space $\reals^d$ to a lower dimensional space of the summary statistics, then the discrepancy could be e.g.~$\discr(\By,\Bx) = ||s(\By) - s(\Bx)||$, where $||\cdot||$ denotes some (possible weighted) norm. Choosing informative summaries and combining them in a reasonable way affect the resulting approximate posterior \citep{Marin2012,Fearnhead2012} but we do not consider this problem here. 

Given $N$ samples from the simulation model with a chosen parameter $\Btheta$, so that $\Bx^{(i)}_{\Btheta} \sim p(\Bx\cond\Btheta),\, i=1,\ldots,N$, the 
ABC posterior at $\Btheta$ can be estimated using 
\begin{align}
p_{\text{ABC}}(\Btheta \cond \By) \approxprop p(\Btheta) \sum_{i = 1}^{N}  \indic_{\discr(\By,\Bx^{(i)}_{\Btheta}) \leq \epsilon}. \label{eq:abc_approx} 
\end{align}
Alternatively, one can use ABC rejection sampling to sample from the ABC posterior, with the following steps: 
1. Draw $\Btheta^{(i)} \sim p(\Btheta)$,
2. Generate $\Bx^{(i)} \sim p(\Bx\cond\Btheta^{(i)})$ from the simulation model, 
3. Accept $\Btheta^{(i)}$ if $\discr(\By,\Bx^{(i)}_{\Btheta}) \leq \epsilon$. 
The accepted values $\{\Btheta^{(i)}\}$ are samples from the approximate posterior distribution. For further background on ABC, we refer the reader to the recent review by \citet{Lintusaari2016}.

\subsection{BOLFI method} \label{subsec:gp-abc}

To speedup inference, \citet{Gutmann2015} proposed to model the discrepancy $\discr_{\Btheta} = \discr(\By,\Bx_{\Btheta})$ between the observed data $\By$ and the simulated data $\Bx_{\Btheta}$ as a function of $\Btheta$. At each step of their algorithm, the current training data i.e.~the discrepancy-parameter pairs $D_{t} = \{(\discr^{(i)},\Btheta^{(i)})\}_{i=1}^{t}$, are used to train the discrepancy model, which is then used to intelligently select the next parameter value $\Btheta^{(t+1)}$ to run the computationally costly simulation model, and thus to obtain updated training data $D_{t+1}$. The simulations can be adaptively focused to areas yielding small discrepancy values (\textit{exploitation}), while allowing some \textit{exploration} of new areas with potentially small values.

At each step, the fitted surrogate model is used to compute an estimated ABC posterior. As opposed to Eq.~\ref{eq:abc_approx}, the estimated posterior for each $\Btheta$ can be obtained as $p(\Btheta) \, \prob(\discr_{\Btheta} \leq \epsilon)$, where the probability is computed using the statistical model (i.e.~the fitted GP). 
For any continuous and strictly increasing function $g$, it holds that $\prob(\discr_{\Btheta} \leq \epsilon) = \prob(g(\discr_{\Btheta}) \leq \epsilon')$, where $\epsilon' = g(\epsilon)$. Thus one can also model $g(\discr_{\Btheta})$ instead to $\discr_{\Btheta}$, which facilitates straightforward transformations for the discrepancy (e.g.~the logarithm) possibly making the discrepancy easier to model.

\subsection{GP models for ABC} \label{subsec:gps}

In this section we describe different GP formulations for modelling the (possibly transformed) discrepancy in the BOLFI approach. In addition to the standard GP model, we include two novel extensions (see below): the input-dependent GP and the classifier GP. We assume that the training data consists of discrepancy-parameter pairs $D_t = \{(\discr^{(i)},\Btheta^{(i)})\}_{i=1}^t$ from the modal area of the posterior, and the aim is to model the discrepancy and the resulting posterior as accurately as possible using $D_t$.

In the \textbf{standard GP regression} one assumes that 
$\discr_{\Btheta} \sim \Normal(f(\Btheta),\sigma^2)$ and $f(\Btheta) \sim \GP(m(\Btheta),k(\Btheta,\Btheta'))$ 
with a mean function $m:\BTheta\rightarrow\reals$ and covariance function $k:\BTheta\times\BTheta\rightarrow\reals$. 
We use $m(\Btheta) = 0$, and the squared exponential covariance function
$k(\Btheta,\Btheta') = \sigma_f^2 \exp\left( -\sum_{i=1}^{p}(\theta_i - \theta'_i\right)^2/(2l_i^2))$ in our experiments.
Given the hyperparameters $\Bphi = (\sigma_f^2, l_1,\ldots,l_p, \sigma^2)$ and training data $D_t$, the posterior predictive density for the latent function $f$ at $\Btheta$ follows a Gaussian density with mean and variance
\begin{align}
\mu_t(\Btheta) &= \Bk_t(\Btheta)^T\BK_t^{-1}(\Btheta)\discr^{(1:t)}, \quad 
v_t(\Btheta) = k(\Btheta,\Btheta) - \Bk_t(\Btheta)^T\BK_t^{-1}(\Btheta)\Bk_t(\Btheta), 
\end{align}
respectively. Above we have denoted $\Bk_t(\Btheta) = (k(\Btheta,\Btheta^{(1)}),\ldots,k(\Btheta,\Btheta^{(t)}))^T$,  $[\BK_t(\Btheta)]_{ij} = k(\Btheta^{(i)},\Btheta^{(j)}) + \sigma^2\indic_{i=j}$ for $i,j=1,\ldots,t$ and $\discr^{(1:t)} = (\discr^{(1)},\ldots,\discr^{(t)})^T$. 
The hyperparameters $\Bphi$ are estimated by maximising the marginal likelihood, for details, see \citet{Rasmussen2006}. 
A model-based estimate of the likelihood at $\Btheta$ can be obtained from the fitted GP as
\begin{align}
\prob(\discr_{\Btheta} \leq \epsilon) = \Phi(({\epsilon - \mu_t(\Btheta))}/{\sqrt{v_t(\Btheta) + \sigma^2}}), \label{eq:gp_post_approx}
\end{align}
where $\epsilon$ is the threshold and $\Phi$ is the cumulative distribution function of the standard Gaussian distribution. An estimate of the posterior density is obtained by multiplying the estimated likelihood with the prior $p(\Btheta)$.

Next we describe the \textbf{input-dependent GP model} \citep{Goldberg1997,Tolvanen2014}. In the standard GP model the noise variance $\sigma^2$ representing the stochasticity in the discrepancy due to simulation is assumed constant. We relax this by assuming  
$\discr_{\Btheta} \sim \Normal(f(\Btheta),\sigma^2\exp(g(\Btheta)))$,  $f(\Btheta) \sim \GP(m(\Btheta),k(\Btheta,\Btheta'))$ and $g(\Btheta) \sim \GP(m_n(\Btheta),k_n(\Btheta,\Btheta'))$. 
That is, also the variance of the discrepancy is modelled with a GP allowing it to change smoothly as a function of the parameter $\Btheta$. Since the variance must be positive, its logarithm is modelled with the GP. As before we set $m(\Btheta) = 0$, and also $m_n(\Btheta) = 0$, implying that a priori the average variance is close to $\sigma^2$. We use the squared exponential covariance functions 
$k(\Btheta,\Btheta') = \sigma_f^2 \exp\left( -\sum_{i=1}^{p}(\theta_i - \theta'_i)^2/(2l_{f_i}^2) \right)$ 
and
$k_n(\Btheta,\Btheta') = \sigma_{g}^2 \exp\left( -\sum_{i=1}^{p}(\theta_i - \theta'_i)^2/(2l_{g_i}^2) \right)$.
There are $2p+2$ hyperparameters to be estimated: $p$ lengthscale parameters, $l_{f_i}$, $l_{g_i}$, and one signal variance parameter for each covariance function, $\sigma_{f}^2$, $\sigma_{g}^2$. The value of $\sigma^2$ is fixed to make the covariance hyperparameters identifiable. Laplace approximation is used for model fitting. We also experimented with the expectation propagation approximation by \citet{Tolvanen2014}, but this came with additional cost and results were qualitatively similar. Eq.~\ref{eq:gp_post_approx} can still be used to estimate the likelihood, by replacing the point estimate of $\sigma^2$ with an estimate of  $\sigma^2\exp(g(\Btheta))$.

The GP models above are used for modelling the ABC discrepancy between observed and simulated data. However, for computing the approximate posterior, it is sufficient to know the probability that the discrepancy is below the threshold $\epsilon$. Motivated by this, we propose a method, \textbf{classifier GP}, which models the lower tail probability directly as a function of the parameter $\Btheta$, using binary GP classification. We interpret the observations $z_i = 2\indic_{\discr^{(i)} \leq \epsilon}-1$ as class labels $+1$ and $-1$ such that 
$p(z_i\cond f(\Btheta_i)) = \lambda^{-1}(z_i f(\Btheta_i))$, where $\lambda$ is either the logit or probit link function and $f(\Btheta) \sim \GP(m(\Btheta),k(\Btheta,\Btheta'))$. 
Hence, this corresponds to an assumption that the discriminative function is smooth, but does not impose additional assumptions about the distribution of the discrepancy. For each parameter value $\Btheta$ the model thus specifies the probability of the discrepancy being classified as +1, i.e., to be below the threshold. The likelihood estimate is thus obtained directly. Unlike with other GP models, we add an additional constant to the prior mean function $m(\Btheta)$ to take into account the fact that the lower tail probabilities are generally very small. Without this, the discriminative function tended to become nonzero near the parameter bounds, inducing posterior mass near the boundaries and, consequently, poor approximations. We use the squared exponential covariance function for the latent function $f$ as for the standard GP method, and Laplace approximation model fitting, see \citet{Rasmussen2006} for details.

\subsection{GP model selection} \label{sec:gps}

Since the distribution of the discrepancy depends on the characteristics of the simulation model and the chosen discrepancy (see e.g.~Table \ref{table:toy_problems} for some potential choices), some GP models will fit the training data better than others. 
Consequently, we propose two utility functions for comparing GP models and different transformations of discrepancy, with the aim of choosing the GP formulation that yields the most accurate estimate of the posterior. See e.g.~\citet{bernardo2001bayesian,vehtari2012} for a thorough discussion on using expected utility for model selection.


As the first criterion, we consider the expected log predictive density for a new discrepancy value $\discr^{(t+h)}$ evaluated at some future evaluation point $\Btheta^{(t+h)}$ for $h=1,2,\ldots$. Here the utility of a single observation $\discr^{(t+h)}$ is defined by 
\begin{align}
u_h =  \log p(\discr^{(t + h)} \cond \Btheta^{(t + h)}, D^{(1:t)}, M),
\end{align}
where $D^{(1:t)} = \{(\discr^{(i)}, \Btheta^{(i)})\}_{i=1}^t$ denotes the training data gathered thus far and $M$ denotes the model. The different transformations of the discrepancy are taken into account by considering the effect of the transformation $\discr' = g(\discr)$. 
The expected utility estimate is obtained by averaging over all the possible realisations of the future data yielding $\bar{u} = \mean_h(u_h)$. 
This utility measures how well the GP predicts the distribution of the discrepancies, which is used for computing the posterior estimate of the simulation model. As we do not know the distribution of the discrepancy-parameter data $(\discr^{(t+h)},\Btheta^{(t+h)})$, we approximate the expected utilities using the data $D^{(1:t)}$ \citep{Vehtari2002}. 
$K$-fold cross-validation (CV) leads to the following estimate for expected log predictive density
\begin{align}
\bar{u}_{\text{CV}} = \frac{1}{t}\sum_{i=1}^t \log p(\discr^{(i)} \cond \Btheta^{(i)}, D^{(1:t)\setminus s(i)}, M),
\end{align}
where the data are split into $K$ (almost) equally sized groups and $s(i)$ denotes the indexes of the group to which the $i$th data point belongs. In practice, we use $K=10$. In the sequel, we refer to this as the \textbf{\mlpd{}}, which stands for the mean of the log-predictive density.


A downside of the mlpd utility is that it does not acknowledge the final purpose of the selected GP model, i.e., to approximate the posterior distribution. 
It may thus give high scores to GP models which broadly model the discrepancy accurately, whereas the focus should be on how well the smallest discrepancies are modelled, as those affect the posterior approximation most. Motivated by this, we frame the problem as a classification task which then leads to a new utility function tailored for ABC inference. The utility for a single observation $\discr^{(t+h)}$ is defined by
\begin{align} 
u_h^c = \indic_{\discr^{(t+h)} \leq \epsilon} \log(\prob(\discr^{(t+h)} \leq \epsilon\cond M)) + \indic_{\discr^{(t+h)} > \epsilon} \log(\prob(\discr^{(t+h)} > \epsilon\cond M)), 
\end{align}
where $\prob(\discr^{(t+h)} \leq \epsilon\cond M)$ is the probability that a new realisation of the discrepancy $\discr^{(t+h)}$ at a test point $\Btheta^{(t+h)}$ is smaller than the threshold $\epsilon$ according to model $M$ (conditioning on $\Btheta^{(t+h)}$ and $D^{(1:t)}$ is omitted to simplify notation). This utility penalises realisations of the discrepancy that are under the threshold when, according to the model, this should happen only with a very small probability, or vice versa. An additional advantage of this utility is that it is invariant to a transformation of the discrepancy if the threshold $\epsilon$ is transformed accordingly, and also it can be used to compare the classifier GP to other models, as it only requires the probability that the discrepancy is below the threshold. 
Again, we use the $K$-fold CV with $K=10$ to approximate the expected utility, so that 
\begin{align}
\bar{u}_{\text{CV}}^c  = & \frac{1}{t}\sum_{i = 1}^t (\indic_{\discr^{(i)} \leq \epsilon} \log(\prob(\discr^{(i)} \leq \epsilon\cond D^{(1:t)\setminus s(i)}, M)) \\
&+ \indic_{\discr^{(i)} > \epsilon} \log(\prob(\discr^{(i)} > \epsilon\cond D^{(1:t)\setminus s(i)}, M))). \nonumber
\end{align}
We call this the \textbf{\cmlpd{}} from now on.

\section{Results} \label{sec:examples}

\subsection{Toy examples} \label{subsec:toy_examples}

We consider several toy examples to study the approximation error for different GP models and transformations of the discrepancy. A summary of the test problems is given in Table \ref{table:toy_problems}. Although simple, these examples highlight potential challenges in modelling the discrepancy that we expect carry over to many realistic problems of potentially higher dimensionality. 
The quality of the results is assessed by computing the total variation distance (TV) between the estimated and the corresponding true posterior i.e.
$\text{TV}(p_{\text{true}},p_{\text{approx}}) = 1/2\int_{\Theta} | p_{\text{true}}(\Btheta) - p_{\text{approx}}(\Btheta) | \ud \Btheta$. Values for this integral are computed numerically. 
Kullback-Leibler divergence (KL), defined as $\text{KL}(p_{\text{true}} \,||\, p_{\text{approx}}) = \int_{\Theta} p_{\text{true}}(\Btheta) \log(p_{\text{true}}(\Btheta)/p_{\text{approx}}(\Btheta)) \ud \Btheta$ is used as an alternative criterion and is also computed numerically. GPstuff 4.6 \citep{Vanhatalo2013} is used for fitting the GP models.

\begin{table} 
\ra{1.2}
\begin{center}
  \begin{tabular}{@{}lp{3.7cm}lp{3.8cm}ll@{}} 
	\hline    
    Test problem and model & prior & $n$ & discrepancy $\discr_{\Btheta}$ & true $\Btheta$ \\ \hline
    Gaussian 1, $\Normal(\theta,1)$ & $ \Unif([-0.5,3])$ & $10$ & $(\bar{y} - \bar{y}({\theta}))^2$ & $1$ \\ 
    Bimodal, $\Normal(\theta^2,2)$ & $\Unif([-2.5,2.5])$ & $5$ & $(\bar{y} - \bar{y}({\theta}))^2$ & $\pm1$ \\ 
    Gaussian 2, $\Normal(0, \theta)$ & $\Unif([0,5])$ & $10$ & $\big(\sigma^2_{y} - \sigma^2_{y({\theta})}\big)^2$ & $1$ \\ 
    Poisson, $\Poi(\theta)$ & $\Unif([0,5]) $ & $10$ & $(\bar{y} - \bar{y}({\theta}))^2$ & $2$ \\ 
    GM 1, $\GM(0.7,\theta,\theta+5,1,2)$ & $\Unif([-10,5])$ & $1$ & $(y_1 - y_{1}({\theta}))^2$ & $1$ \\ 
    GM 2, $\GM(0.7,\theta,\theta,3,0.25)$ & $\Unif([-6,6])$ & $1$ & $(y_1 - y_{1}({\theta}))^2$ & $1$ \\ 
    Uniform, $\Unif([0,\theta])$ & $\Unif([0,5])$ 
    & $5$ & $(\max\{y_i\} - \max\{y_{i}(\theta)\})^2$ & $2$ \\ 
    2D Gaussian 1, $\Normal(\Btheta,\BSigma)$ & $\Unif([1.5,4]\times[1.5,4])$ & $10$ & $(\bar{\By} - \bar{\By}({\theta}))^T\BSigma^{-1}(\bar{\By} - \bar{\By}({\theta}))$ & $[2.5, 2.5]^T$ \\ 
    2D Gaussian 2, $\Normal(\theta_1,\theta_2)$ & $\Unif([2,4.5]\times[0.5, 5])$ & $25$ & $(\bar{y} - \bar{y}({\theta}))^2 + \big(\sigma^2_{y} - \sigma^2_{y({\theta})}\big)^2$ & $[3, 2]^T$ \\ 
    Lotka-Volterra, see the text & $\Unif([0.25,1.25]\times[0.5,1.5])$ & $8$ & see the text & $[1, 1]^T$ \\ \hline
  \end{tabular}
  \caption[This text causes matrix not to produce errors in caption...]{
  Description of the test problems. Above, $\bar{y}$ denotes the sample mean of $\{y_i\}_{i=1}^n$ and, similarly, $\sigma^2_{y}$ is the sample variance. The data points $y_i(\theta)$ are independent and identically distributed draws from the simulation model with parameter $\theta$. Also, $\GM(\alpha,\mu_1,\mu_2,\sigma_1^2,\sigma_2^2) = \alpha\Normal(\mu_1,\sigma_1^2) + (1-\alpha)\Normal(\mu_2,\sigma_2^2)$. For the 2D Gaussian we use a fixed covariance matrix $\BSigma$ with unit variances and correlation $0.5$. 
    } \label{table:toy_problems}
\end{center}
\end{table}

We consider two transformations of the squared error (\textit{se}) discrepancy shown in Table \ref{table:toy_problems}, namely, the log and the square-root transformations (\textit{log} and \textit{sqrt}). Although other transformations can be used, these already demonstrate the main findings. 
One could also transform the individual summaries before combining them to a discrepancy function, but we do not consider this approach here. 
We use uniform priors for the parameters of the simulation models over a range covering the modal area of the true posterior. We also repeat the experiments with a much wider support, although this seems less relevant in practice when the goal is to obtain an accurate posterior estimate where the majority of mass is located, and hence focus simulations there. Other priors and adaptive schemes for choosing the training data are also possible \citep{Gutmann2015}. 
We set the ABC threshold $\epsilon$ customarily as the $0.05$th quantile of the discrepancies sampled from the uniform prior and use the same threshold for all GP-ABC methods, the baseline ABC rejection sampler, and for the ``true'' ABC posterior computed using ABC rejection sampling with extensive simulations. Thus the difference in results is only caused by the choice of GP model and discrepancy transformation. 

To get an estimate of the variability due to a stochastic simulation model, we repeat each experiment $100$ times. 
We also repeat the experiments with some other choices of the threshold and using the true posterior density (which is available analytically for the test problems) as the baseline. These results are presented in the appendices \ref{subsec:post_est_truepost} and \ref{subsec:threshold}.
For the basic ABC rejection sampler, we use kernel density estimation as a post-processing step to approximate the posterior curve from the accepted samples, thereby imposing basic smoothness assumptions of the posterior. 
We use the logit link function for GP classifier method in our experiments.
The full summary of the results is gathered in Tables \ref{table:all_results} and \ref{table:all_results2d}, and below we analyse in detail some of the key findings.


\begin{example}[Gaussian 1]
As the first example, representing many key findings, we consider a simple Gaussian model with an unknown mean and known variance, see ``Gaussian 1'' in Table \ref{table:toy_problems}. This model is simple enough to be analysed analytically. Consider the discrepancies $\discr_{\theta} = (\bar{y} - \bar{x}_{\theta})^2$ and $\discr_{\theta}' = |\bar{y} - \bar{x}_{\theta}|$. Using basic properties of the expectation and the Gaussian distribution, we obtain $\mean(\discr_{\theta}) = (\theta - \bar{y})^2 + \sigma^2/n$ and $\var(\discr_{\theta}) = 2\sigma^2(2(\theta - \bar{y})^2 + 1)/n$.
Similarly $\var(\discr_{\theta}') \approx \sigma^2/n$, which holds accurately for large $|\theta|$. 
We see that the variance of the discrepancy $\discr_{\theta}$ grows quadratically as a function of the parameter $\theta$. On the other hand, with $\discr_{\theta}'$ the variance is approximately constant.

The main observations of this example are illustrated in Figure \ref{fig:demo_gaussian}. In (a-c) a prior over a wide range is used and in (d-f) training data are gathered within a narrower region around the mode. Comparing (a) and (b) shows that the input-dependent GP model yields a much better approximation than the standard GP. In (a) the poor GP fit causes also a poor approximation to the posterior, which cannot be corrected by increasing the number of simulations. Furthermore, different transformations change the behaviour of the discrepancy. The square-root -transformation in (c), which makes the variance of the discrepancy approximately constant, improves the standard GP considerably. In (d-f) three different GP models are fitted near the posterior mode, and we see that focusing the simulations to the central region improves the performance of all methods.

\begin{figure}[ht]
\centering
\includegraphics[width=0.9\textwidth]{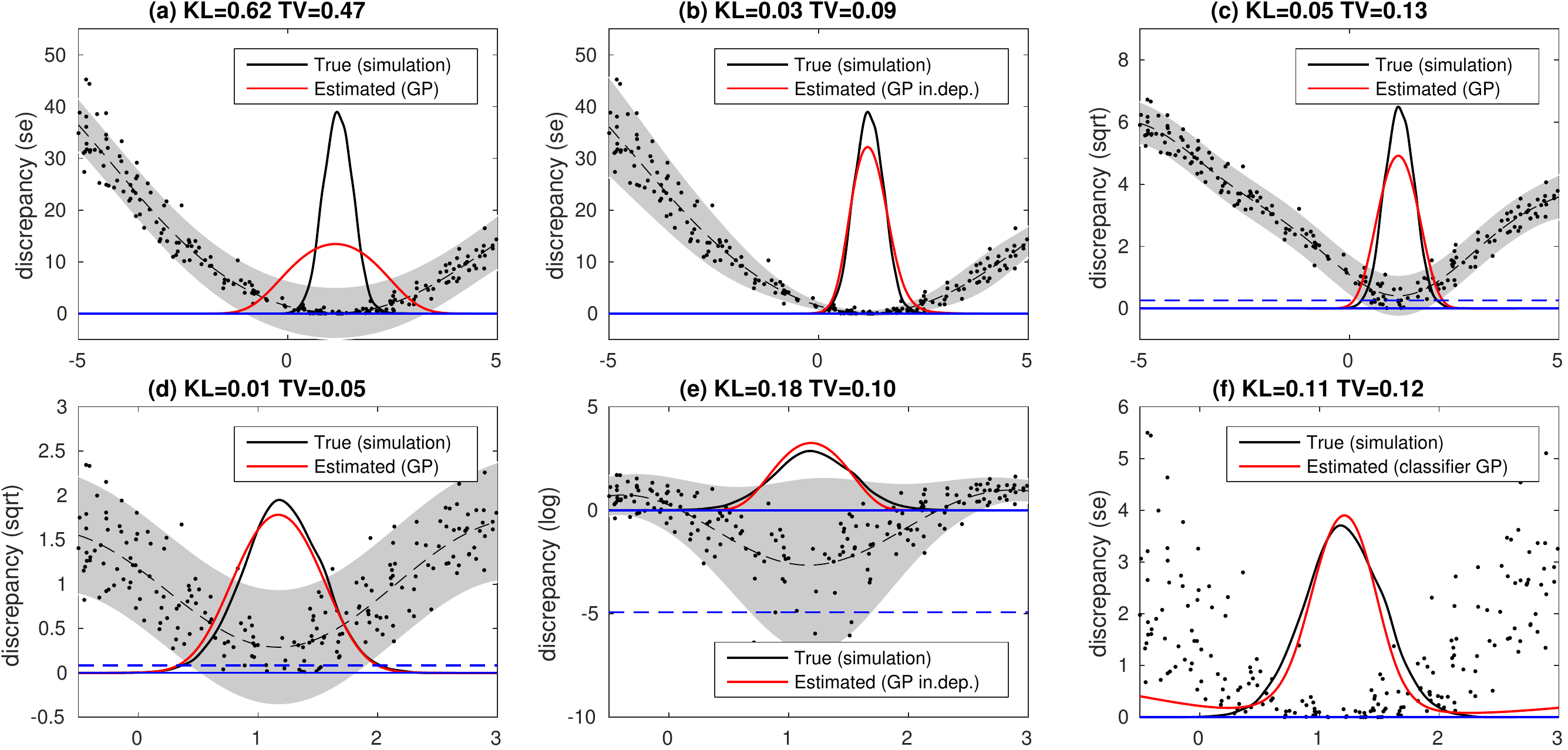}
\caption
{Results for the ``Gaussian 1'' model. The grey area is the $95\%$ probability interval, blue dashed line is the threshold and the black dots represent realisations of the discrepancy. The abbreviations ``se'', ``log'' and ``sqrt'' refer to squared, log transformed and square-root transformed discrepancies, respectively. In (a) the GP is fitted to discrepancy realisations on a wide interval resulting in a poor approximation. Better approximations are obtained by the input-dependent GP model (b) or transforming the discrepancy (c). In (d) the fitting is done on the area of significant posterior mass resulting in the best fit in terms of both TV and KL, even if the variance of the discrepancy is still clearly overestimated in the modal region. In (e) the posterior uncertainty is slightly underestimated due to the skewness of the log-transformed discrepancy. The classifier GP in (f) slightly overestimates the tails of the posterior. 
} \label{fig:demo_gaussian}
\end{figure}
\end{example}


\begin{example}[Poisson] 
We estimate the parameter of the Poisson distribution which demonstrates the benefit of GP modelling compared to the ABC rejection sampling. Figure \ref{fig:demo_poi} shows typical results. Here the data have discrete values but the discrepancy is approximately Gaussian, and the variance of the discrepancy grows as a function of the parameter. The input-dependent model does not improve the results visibly, even if the fit to the discrepancy data is evidently better. 
The best approximations are obtained when the square-root transformation is used as in (a-b) since then the discrepancy is approximately Gaussian, although its variance is not constant. The ABC rejection sampler in (c) does not work well due to the small number of accepted samples as compared to the GP-based methods.

\begin{figure}[ht]
\centering
\includegraphics[width=0.9\textwidth]{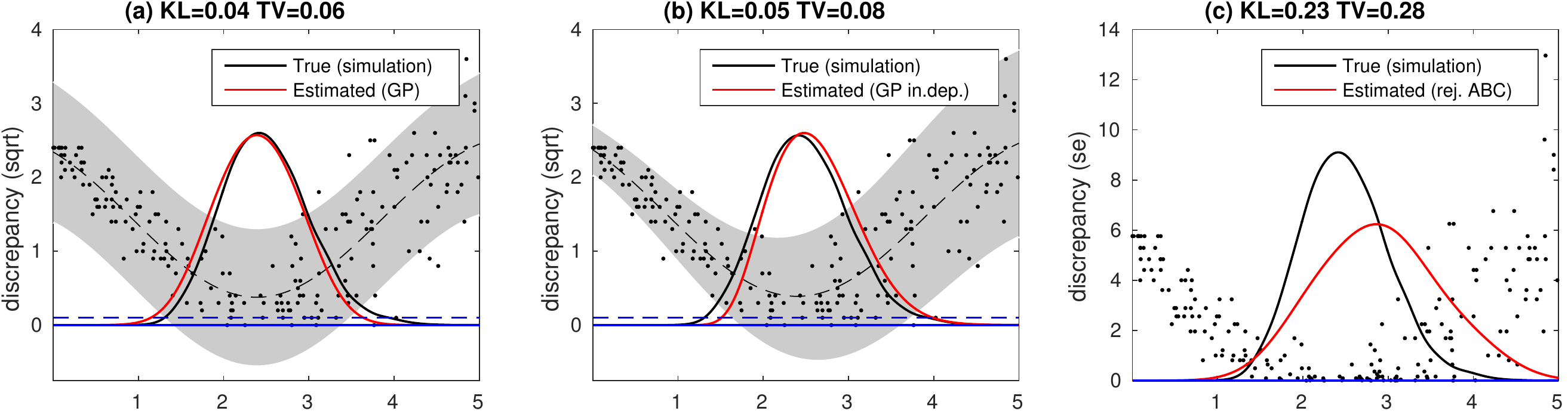}
\caption
{Results for the ``Poisson'' example, demonstrating the benefits of the GP modelling. The input-dependent GP model (b) fits the discrepancy data better than the standard GP (a). Despite this difference, the posterior approximations are about equally good. The ABC rejection sampler (c) yields a less accurate approximation with only the $200$ training points available.
} \label{fig:demo_poi}
\end{figure}
\end{example}


\begin{example}[GM 1] \label{ex:bimodal}
The third example demonstrates that both the standard and input-dependent GP models may fail to capture a bimodal shape of the posterior. Here also the discrepancy distribution is bimodal conditional on specific parameter values. A particular realisation is shown in Figure \ref{fig:demo_gm}. The GP and input-dependent GP yield slightly different approximations, neither of which captures the bimodality. However, fixing the lengthscale to a small value allows to capture the bimodal shape but with the cost of making the overall shape of the estimated posterior wiggly (not shown). The ABC rejection sampler works better despite the limited training set size of $200$. 
This observation does not hold for all bimodal posteriors, though. To demonstrate this, we designed another example where the posterior is bimodal, the ``Bimodal'' in Table \ref{table:toy_problems}. In contrast with the Gaussian mixture model above, the distribution of the discrepancy is close to a Gaussian for any parameter. This type of discrepancy can be modelled well, and consequently, the bimodal shape of the posterior can be learnt accurately.

\begin{figure}[ht]
\centering
\includegraphics[width=0.9\textwidth]{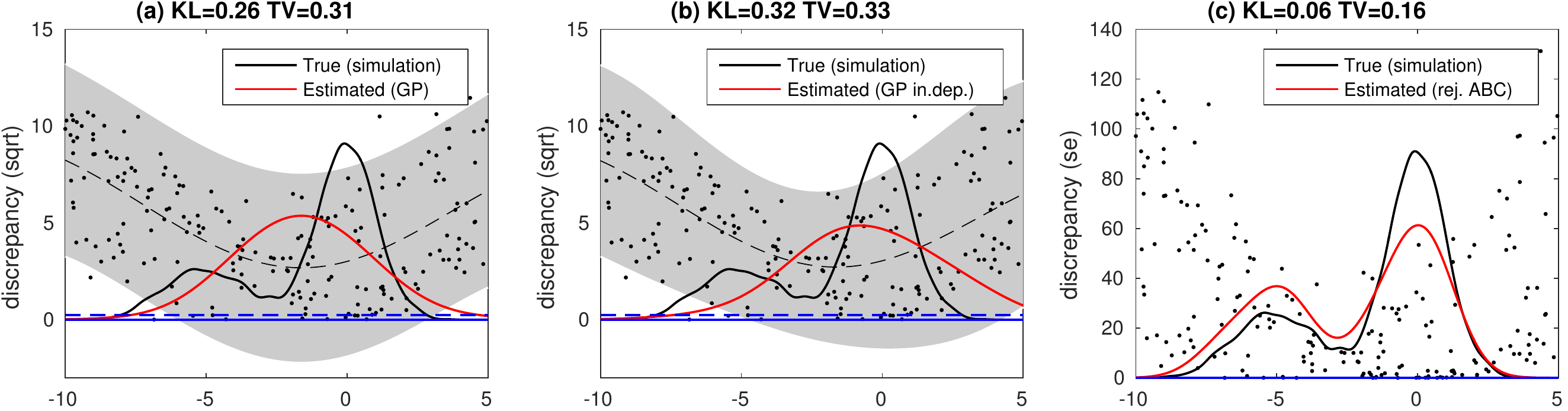}
\caption
{Neither the standard (a) nor the input-dependent GP model (b) learn the shape of the posterior in the bimodal ``GM 1'' example. Notably, not only the posterior, but also the discrepancy distribution, given a particular parameter value, is bimodal, which is the explanation of this behaviour. $200$ points were generated from the model, but similar results were obtained with a larger set of simulations, and with other transformations. On the other hand, the ABC rejection sampler (c) uncovers the bimodal shape.
} \label{fig:demo_gm}
\end{figure}
\end{example}


\begin{example}[Lotka-Volterra] \label{ex:lv} 
We consider the Lotka-Volterra model used by \citet{Toni2009} to compare ABC methods. The model describes the evolution of prey and predator populations, defined by differential equations
\begin{align}
\frac{\ud x_1}{\ud t} = \theta_1 x_1 - x_1 x_2, \quad \frac{\ud x_2}{\ud t} = \theta_2 x_1 x_2 - x_2,
\end{align}
where $x_1=x_1(t)$ and $x_2=x_2(t)$ describe the prey and predator species at time $t$, respectively. Their initial values are set to $x_1(0) = 0.5$ and $x_2(0) = 1.0$. Vector $\Btheta = (\theta_1,\theta_2)$ is the parameter to be estimated. The $8$ measurements for $(x_1,x_2)$ are corrupted by additive independent and identically distributed Gaussian noise $\Normal(0,0.5^2)$. We consider a discrepancy 
\begin{align}
\discr_{\Btheta} = \sum_{i=1}^{8}\sum_{j=1}^{2}(x_j(t_i) - \hat{x}_j(t_i,\Btheta))^2,
\end{align}
where $x_j(t_i)$ are the noisy measurements at time $t_i$ and $\hat{x}_j(t_i,\Btheta)$ the corresponding predictions with parameter $\Btheta$. The prior and the true value of the parameter vector are shown in Table \ref{table:toy_problems}.

The results are shown in Figure \ref{fig:demo_lv}, and we see that the GP formulation has only a moderate impact. However, the classifier GP and the ABC rejection sampler perform worse than the GP-based methods, as seen also in Table \ref{table:all_results2d}. The estimates of the ABC rejection sampler also vary more between the different simulated training data.

\begin{figure}[h!]
\centering
\includegraphics[width=0.8\textwidth]{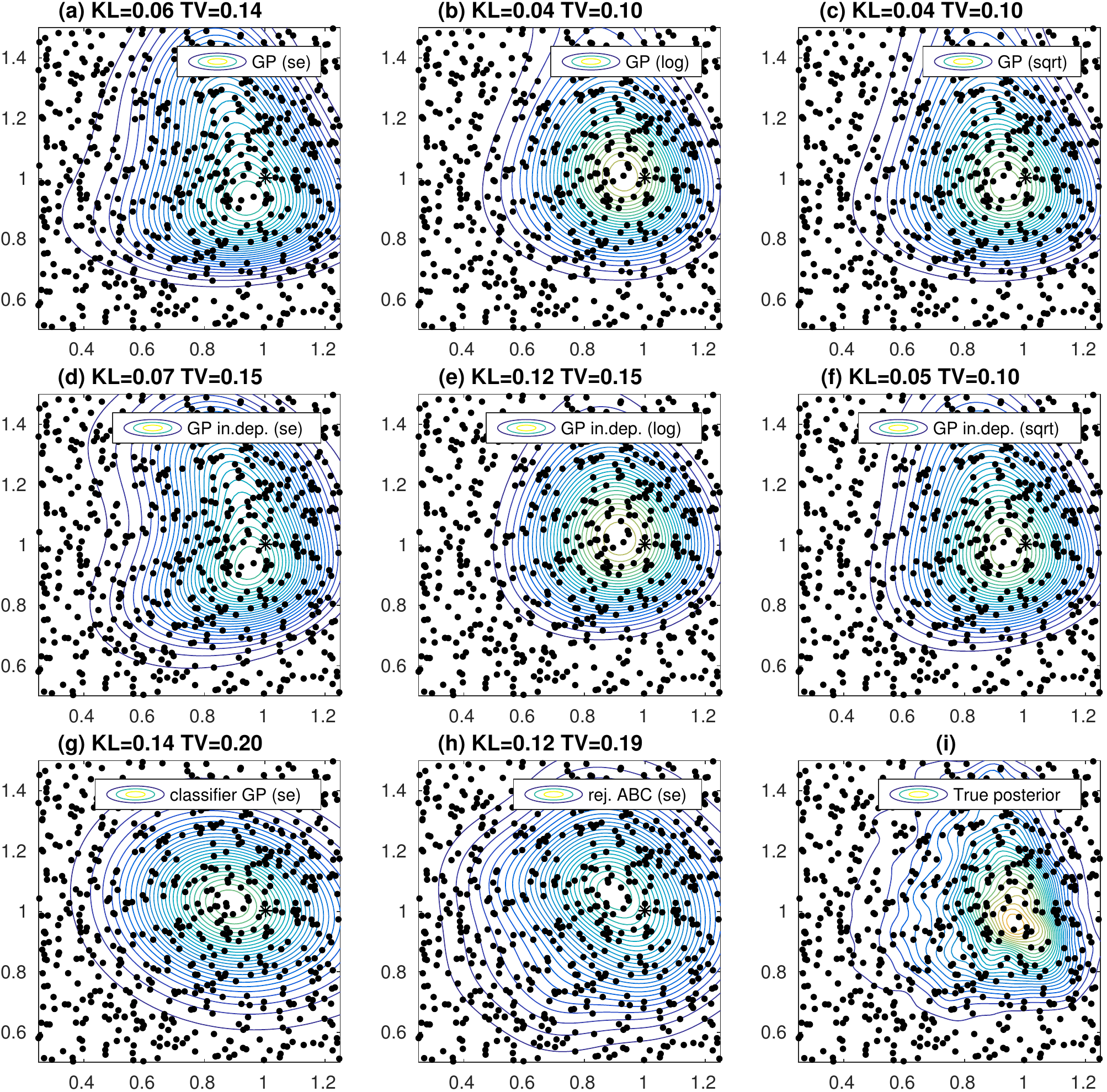}
\caption
{Posterior estimates for the Lotka-Volterra model. The black dots represent points where the simulation was run. 
The parameter $\theta_1$ is on the x-axis and $\theta_2$ on the y-axis. 
The difference between the standard and input-dependent GP formulations is minor, but both of them outperform the classifier GP and the ABC rejection sampler. 
} \label{fig:demo_lv}
\end{figure}
\end{example}


\begin{table} 
\ra{1.1}
\setlength{\tabcolsep}{4.0pt}
\scriptsize 
\begin{center}
\begin{tabular}{@{}llllllllllllllll@{}} 
\toprule
& \multicolumn{3}{c}{n=50} & \multicolumn{3}{c}{n=100} & \multicolumn{3}{c}{n=200} & \multicolumn{3}{c}{n=400} & \multicolumn{3}{c}{n=600} \\ 
\cmidrule{2-4} \cmidrule{5-7} \cmidrule{8-10} \cmidrule{11-13} \cmidrule{14-16} 
& se & log & sqrt & se & log & sqrt & se & log & sqrt & se & log & sqrt & se & log & sqrt \\ \midrule
Gaussian 1:\\
GP & 0.17 & 0.09 & \textbf{0.07} & 0.20 & 0.10 & \textbf{0.05} & 0.21 & 0.11 & \textbf{0.04} & 0.20 & 0.17 & \textbf{0.03} & 0.20 & 0.18 & \textbf{0.03} \\
GP in.dep. & 0.19 & 0.14 & 0.09 & 0.18 & 0.14 & 0.06 & 0.18 & 0.13 & 0.05 & 0.19 & 0.14 & 0.04 & 0.21 & 0.15 & 0.04 \\
\classgp{} & 0.40 & 0.40 & 0.40 & 0.33 & 0.33 & 0.33 & 0.19 & 0.19 & 0.19 & 0.10 & 0.10 & 0.10 & 0.09 & 0.09 & 0.09 \\
\abcrej{} & 0.31 & 0.31 & 0.31 & 0.26 & 0.26 & 0.26 & 0.18 & 0.18 & 0.18 & 0.12 & 0.12 & 0.12 & 0.11 & 0.11 & 0.11 \\ \hline
Bimodal:\\
GP & 0.20 & 0.47 & \textbf{0.16} & 0.20 & 0.26 & \textbf{0.12} & 0.21 & 0.18 & \textbf{0.10} & 0.21 & 0.16 & \textbf{0.08} & 0.20 & 0.17 & \textbf{0.07} \\
GP in.dep. & 0.19 & 0.58 & 0.18 & 0.17 & 0.39 & 0.14 & 0.20 & 0.28 & 0.11 & 0.20 & 0.25 & 0.09 & 0.20 & 0.23 & 0.08 \\
\classgp{} & 0.39 & 0.39 & 0.39 & 0.35 & 0.35 & 0.35 & 0.24 & 0.24 & 0.24 & 0.17 & 0.17 & 0.17 & 0.14 & 0.14 & 0.14 \\
\abcrej{} & 0.45 & 0.45 & 0.45 & 0.39 & 0.39 & 0.39 & 0.26 & 0.26 & 0.26 & 0.20 & 0.20 & 0.20 & 0.16 & 0.16 & 0.16 \\ \hline
Gaussian 2:\\
GP & 0.32 & 0.36 & \textbf{0.27} & 0.32 & \textbf{0.24} & 0.26 & 0.33 & \textbf{0.20} & 0.25 & 0.33 & 0.21 & 0.24 & 0.32 & 0.22 & 0.23 \\
GP in.dep. & 0.48 & 0.37 & 0.30 & 0.45 & 0.33 & 0.29 & 0.44 & 0.29 & 0.27 & 0.42 & 0.25 & 0.26 & 0.41 & 0.25 & 0.25 \\
\classgp{} & 0.35 & 0.35 & 0.35 & 0.35 & 0.35 & 0.35 & 0.26 & 0.26 & 0.26 & \textbf{0.17} & \textbf{0.17} & \textbf{0.17} & \textbf{0.15} & \textbf{0.15} & \textbf{0.15} \\
\abcrej{} & 0.29 & 0.29 & 0.29 & 0.29 & 0.29 & 0.29 & 0.22 & 0.22 & 0.22 & 0.18 & 0.18 & 0.18 & 0.16 & 0.16 & 0.16 \\ \hline
GM 1:\\
GP & 0.32 & \textbf{0.29} & 0.31 & 0.31 & \textbf{0.28} & 0.30 & 0.31 & 0.28 & 0.30 & 0.30 & 0.23 & 0.29 & 0.29 & 0.19 & 0.28 \\
GP in.dep. & 0.42 & 0.35 & 0.32 & 0.40 & 0.37 & 0.31 & 0.39 & 0.34 & 0.30 & 0.38 & 0.25 & 0.29 & 0.37 & 0.24 & 0.28 \\
\classgp{} & 0.34 & 0.34 & 0.34 & 0.36 & 0.36 & 0.36 & 0.27 & 0.27 & 0.27 & \textbf{0.18} & \textbf{0.18} & \textbf{0.18} & \textbf{0.14} & \textbf{0.14} & \textbf{0.14} \\
\abcrej{} & 0.35 & 0.35 & 0.35 & 0.32 & 0.32 & 0.32 & \textbf{0.24} & \textbf{0.24} & \textbf{0.24} & 0.19 & 0.19 & 0.19 & \textbf{0.14} & \textbf{0.14} & \textbf{0.14} \\ \hline
GM 2:\\
GP & 0.19 & 0.14 & \textbf{0.13} & 0.18 & \textbf{0.12} & \textbf{0.12} & 0.19 & 0.12 & \textbf{0.11} & 0.19 & 0.11 & 0.11 & 0.20 & 0.11 & 0.11 \\
GP in.dep. & 0.26 & 0.19 & 0.14 & 0.24 & 0.18 & \textbf{0.12} & 0.21 & 0.16 & \textbf{0.11} & 0.22 & 0.16 & \textbf{0.10} & 0.24 & 0.15 & \textbf{0.10} \\
\classgp{} & 0.37 & 0.37 & 0.37 & 0.33 & 0.33 & 0.33 & 0.21 & 0.21 & 0.21 & 0.14 & 0.14 & 0.14 & 0.12 & 0.12 & 0.12 \\
\abcrej{} & 0.33 & 0.33 & 0.33 & 0.29 & 0.29 & 0.29 & 0.20 & 0.20 & 0.20 & 0.16 & 0.16 & 0.16 & 0.14 & 0.14 & 0.14 \\ \hline
Uniform\\
GP & 0.26 & 0.22 & \textbf{0.15} & 0.26 & 0.24 & 0.15 & 0.27 & 0.22 & 0.15 & 0.26 & 0.23 & 0.15 & 0.26 & 0.23 & 0.15 \\
GP in.dep. & 0.26 & 0.22 & 0.16 & 0.22 & 0.21 & \textbf{0.13} & 0.19 & 0.23 & \textbf{0.12} & 0.17 & 0.23 & \textbf{0.11} & 0.15 & 0.23 & 0.12 \\
\classgp{} & 0.43 & 0.43 & 0.43 & 0.34 & 0.34 & 0.34 & 0.23 & 0.23 & 0.23 & 0.14 & 0.14 & 0.14 & \textbf{0.11} & \textbf{0.11} & \textbf{0.11} \\
\abcrej{} & 0.33 & 0.33 & 0.33 & 0.31 & 0.31 & 0.31 & 0.23 & 0.23 & 0.23 & 0.19 & 0.19 & 0.19 & 0.17 & 0.17 & 0.17 \\ \hline
Poisson:\\
GP & 0.19 & 0.12 & \textbf{0.09} & 0.18 & 0.10 & \textbf{0.07} & 0.18 & 0.08 & \textbf{0.06} & 0.20 & 0.11 & \textbf{0.06} & 0.20 & 0.12 & \textbf{0.06} \\
GP in.dep. & 0.21 & 0.23 & 0.13 & 0.21 & 0.19 & 0.10 & 0.24 & 0.18 & 0.08 & 0.23 & 0.15 & 0.07 & 0.24 & 0.16 & 0.07 \\
\classgp{} & 0.33 & 0.33 & 0.33 & 0.28 & 0.28 & 0.28 & 0.14 & 0.14 & 0.14 & 0.09 & 0.09 & 0.09 & 0.09 & 0.09 & 0.09 \\
\abcrej{} & 0.26 & 0.26 & 0.26 & 0.23 & 0.23 & 0.23 & 0.16 & 0.16 & 0.16 & 0.11 & 0.11 & 0.11 & 0.10 & 0.10 & 0.10 \\ 
\bottomrule
\end{tabular}
\caption{Results for the 1D toy examples. The quality of the approximation was measured using the TV distance between the estimated and the true ABC posterior densities. The smallest TV values are bolded. Value $n$ is the number of model simulations and ``se'', ``log'' and  ``sqrt'' refer to squared, log transformed and square-root transformed discrepancies, respectively.
} \label{table:all_results}
\end{center}
\end{table}

\begin{table} 
\ra{1.1}
\setlength{\tabcolsep}{4.0pt}
\scriptsize 
\begin{center}
\begin{tabular}{@{}llllllllllllllll@{}} 
\toprule
& \multicolumn{3}{c}{n=100} & \multicolumn{3}{c}{n=200} & \multicolumn{3}{c}{n=400} & \multicolumn{3}{c}{n=600} & \multicolumn{3}{c}{n=800} \\ 
\cmidrule{2-4} \cmidrule{5-7} \cmidrule{8-10} \cmidrule{11-13} \cmidrule{14-16}
& se & log & sqrt & se & log & sqrt & se & log & sqrt & se & log & sqrt & se & log & sqrt \\ \midrule
2D Gaussian 1:\\
GP & 0.24 & 0.15 & \textbf{0.12} & 0.23 & 0.13 & \textbf{0.09} & 0.22 & 0.12 & \textbf{0.07} & 0.22 & 0.12 & \textbf{0.07} & 0.22 & 0.12 & \textbf{0.07} \\
GP in.dep. & 0.20 & 0.20 & 0.14 & 0.19 & 0.15 & 0.10 & 0.17 & 0.12 & 0.08 & 0.18 & 0.12 & \textbf{0.07} & 0.19 & 0.11 & \textbf{0.07} \\
\classgp{} & 0.62 & 0.62 & 0.62 & 0.63 & 0.63 & 0.63 & 0.24 & 0.24 & 0.24 & 0.15 & 0.15 & 0.15 & 0.12 & 0.12 & 0.12 \\
\abcrej{} & 0.35 & 0.35 & 0.35 & 0.27 & 0.27 & 0.27 & 0.22 & 0.22 & 0.22 & 0.19 & 0.19 & 0.19 & 0.17 & 0.17 & 0.17 \\ \hline
2D Gaussian 2:\\
GP & 0.53 & 0.32 & 0.45 & 0.51 & 0.26 & 0.43 & 0.51 & 0.22 & 0.40 & 0.50 & 0.21 & 0.39 & 0.50 & 0.20 & 0.39 \\
GP in.dep. & 0.64 & 0.27 & \textbf{0.26} & 0.61 & \textbf{0.20} & 0.24 & 0.50 & \textbf{0.17} & 0.22 & 0.49 & \textbf{0.15} & 0.22 & 0.50 & \textbf{0.13} & 0.21 \\
\classgp{} & 0.63 & 0.63 & 0.63 & 0.63 & 0.63 & 0.63 & 0.36 & 0.36 & 0.36 & 0.21 & 0.21 & 0.21 & 0.17 & 0.17 & 0.17 \\
\abcrej{} & 0.34 & 0.34 & 0.34 & 0.29 & 0.29 & 0.29 & 0.25 & 0.25 & 0.25 & 0.20 & 0.20 & 0.20 & 0.19 & 0.19 & 0.19 \\ \hline
Lotka-Volterra:\\
GP & 0.22 & \textbf{0.19} & 0.20 & 0.18 & \textbf{0.15} & 0.16 & 0.16 & \textbf{0.13} & 0.15 & 0.15 & \textbf{0.12} & 0.13 & 0.15 & \textbf{0.12} & \textbf{0.12} \\
GP in.dep. & 0.31 & 0.23 & 0.23 & 0.21 & 0.19 & 0.19 & 0.18 & 0.16 & 0.15 & 0.16 & 0.15 & 0.14 & 0.15 & 0.14 & \textbf{0.12} \\
\classgp{} & 0.51 & 0.51 & 0.51 & 0.51 & 0.51 & 0.51 & 0.50 & 0.50 & 0.50 & 0.20 & 0.20 & 0.20 & 0.17 & 0.17 & 0.17 \\
\abcrej{} & 0.33 & 0.33 & 0.33 & 0.26 & 0.26 & 0.26 & 0.22 & 0.22 & 0.22 & 0.20 & 0.20 & 0.20 & 0.18 & 0.18 & 0.18 \\ 
\bottomrule
\end{tabular}
\caption{Results for the 2D toy examples. See the caption for Table \ref{table:all_results} for details.} \label{table:all_results2d}
\end{center}
\end{table}


As a general observation from Tables \ref{table:all_results} and \ref{table:all_results2d}, we conclude that whenever the discrepancy is close to a Gaussian, or if the number of evaluations is very small and only a few discrepancy values are below the threshold $\epsilon$, the GP-based approaches yield better posterior approximations than the ABC rejection sampler or the classifier GP method. However, if the Gaussian assumptions are violated, as in Example \ref{ex:bimodal}, the rejection sampler and the classifier GP are more accurate. Increasing the number of simulations does not help as it does not solve the model misspecification. 
Interestingly, as few as $50$ model evaluations in 1D ($200$ in 2D) result in almost as accurate results as $400$ evaluations in 1D ($600$ in 2D). On the other hand, the accuracies of the ABC rejection sampler and classifier GP clearly improve as the number of evaluation points is increased. Additional evaluations also improve the stability of GP estimation and, hence, decrease the variance in the results.

The classifier GP performs generally similarly or slightly better than the ABC rejection sampler. 
However, with a small number of evaluations the error of the classifier GP is relatively large, but as the number of evaluations increases, the accuracy increases rapidly reaching and finally clearly outperforming the ABC rejection sampler. However, decreasing the threshold to the $0.01$th quantile leads to conservative results since the number of realisations of the discrepancy below the threshold becomes very small as shown in appendix. 
Typically the classifier GP tends to overestimate the probability in the posterior tail area despite our attempts to change this behaviour as described in Section \ref{subsec:gps}.

Overall, the square-root transformation seems to work best while log-transformation is also useful in some cases. 
However, with small threshold values, such as the $0.01$th quantile of the realised discrepancies, modelling the log-transformed discrepancy with a GP tends to cause too narrow posterior distributions in some scenarios, see Figure \ref{fig:demo_gaussian}(e). This happens if many discrepancies are close to zero, in which case the log transformation results in a strongly skewed distribution. This may not be an issue in practice, since with a complex model simulating data such that the discrepancy becomes very small is unlikely (or impossible if the model is misspecified), even with the optimal parameter value. 
Also, modelling non-negative discrepancies with GP regression does not appear to cause large additional posterior approximation error in practice, see e.g.~Figures \ref{fig:demo_gaussian} and \ref{fig:demo_poi}. In some of the test cases, the input-dependent GP model worked best but a similar effect was often achieved also by modelling a suitably transformed discrepancy.


\subsection{Model selection results} \label{subsec:model_comp_examples}

In Section \ref{sec:gps}, we formulated two utility functions to guide the selection of the GP model: the expected log predictive density (\mlpd{}) and the expected log predictive probability of attaining a discrepancy that falls below the threshold (\cmlpd{}). Next we illustrate the performance of these methods in practice. 
We consider the same toy problems as in Section \ref{subsec:toy_examples}, and we exclude the classifier GP from the comparisons related to the \mlpd{}. 

\begin{figure}[ht]
\centering
\includegraphics[width=.9\textwidth]{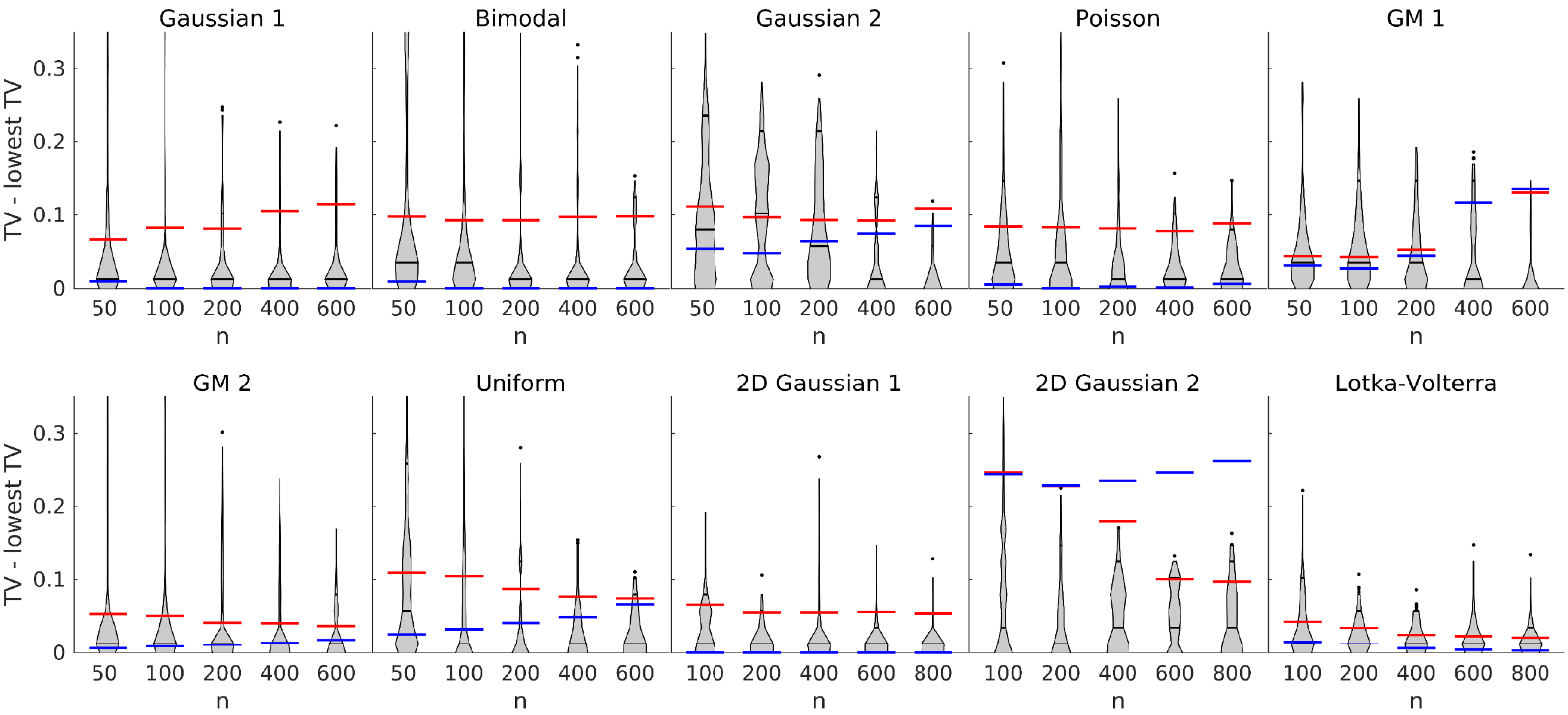}
\caption
{Results of the GP model selection using the \cmlpd{}. 
The value on the y-axis is the difference between the TV distance of the chosen GP (corresponding to the largest utility) and the smallest TV distance observed (corresponding to the most accurate result obtained). Therefore, the smaller the value is, the closer the selected model is to the optimal model. The violin plot shows the results over 100 simulated training data sets. The x-axis shows the number of model simulations $n$. The blue line represents the median results if the standard GP with the square-root transformation is always chosen. Another baseline shown with red is obtained by randomly selecting the GP model formulation.
} \label{fig:mod_sel_our_logscore}
\end{figure}

The results in Figure \ref{fig:mod_sel_our_logscore} and Figure \ref{fig:mod_sel_gp_mlpd} in appendix \ref{subsec:gp_model_sel}
demonstrate the performance of the mlpd and classifier utilities, when used to select a GP model to estimate the posterior. We see that both methods work reasonably well across all cases, although ``Gaussian 2'' and ``2D Gaussian 1'' toy problems seem more difficult than the rest. Also, as expected, as more simulations become available, the model selection improves in most scenarios, such that the highest utilities better identify the GP formulations resulting in the most accurate posterior approximations. 
For some individual simulations a GP model resulting in a poor posterior approximation has the highest utility. This happens mainly with a small number of simulations and the \cmlpd{}, because then the number of cases below the threshold is very small, and, consequently, the utility itself has a high variance. These cases are seen as peaks in the violin plots.

Comparison of Figure \ref{fig:mod_sel_our_logscore} and Figure \ref{fig:mod_sel_gp_mlpd} shows that the overall performance difference between the two proposed utilities is relatively small. 
However, in the case of ``Gaussian 2'' and ``GM 1'' examples, the \mlpd{} performs systematically worse than the \cmlpd{}. In these cases even with $600$ evaluations, the \mlpd{} tends to propose suboptimal GP models. Further, the \cmlpd{} can be used to compare basically any set of models that predict the amount of posterior mass under the threshold, making it more applicable than the \mlpd{} as explained in Section \ref{sec:gps}. On the other hand, the performance of the \cmlpd{} criterion is more dependent on the value of the threshold. When the threshold is decreased so that only a few discrepancies fall below the threshold, the method will not work anymore, contrary to the \mlpd{}.


\subsection{Horizontal gene transfer between bacterial genomes} \label{subsec:real_examples}

The emerging field of bacterial genomics involves analysis of thousands of bacterial genomes, to understand the variability in bacteria as well as to answer questions of practical importance, such as the spread of antibiotic resistance \citep{croucher2011rapid,chewapreecha2014dense}. One interesting observation is the extent to which members of the same bacterial species can differ in genome content, i.e., different strains of the same species can have different sets of genes, and only a minority of the genes is observed in all strains \citep{touchon2009organised}. Furthermore, bacteria can exchange genes with one another in a process called horizontal gene transfer (HGT) \citep{thomas2005mechanisms}.

Here we consider a previously published population genomic model that describes the variation in genome content \citep{Marttinen2015}. Point estimates of the parameters have previously been published for this model, but we are interested in estimating the full posterior, when the model is fitted to a published collection of 616 genomes from \textit{Streptococcus pneumoniae} \citep{croucher2013population}. Briefly, the model consists of a forward-simulation of a population of bacterial strains for many generations. At each generation, the next generation is simulated by selecting strains randomly from the current generation. In addition, the genome content of the descendants may be modified by three operations, the rates of which correspond to the three parameters of the model: the gene deletion rate (\textit{del}), novel gene introduction rate (\textit{nov}), and the rate of HGT where the gene presence-absence status of the donor strain is copied to the recipient strain (\textit{hgt}).

To estimate the model parameters, we consider the discrepancy 
\begin{align}
\discr_{\Btheta} = w_1 \text{KL}(\Btheta) + w_2 (c_{\text{real}} - c_{\text{simu}}(\Btheta))^2,\label{eq:real_discrepancy}
\end{align}
where $\text{KL}(\Btheta)$ is the Kullback-Leibler divergence between the observed and simulated gene frequency spectra, $c_{\text{real}}$ is the so-called observed clonality score, and $c_{\text{simu}}(\Btheta)$ the corresponding simulated value, see \citet{Marttinen2015} for details. The weights $w_1$ and $w_2$ are used to transform the summaries approximately on the same scale, which is common in ABC literature. 
\citet{Marttinen2015} achieved the same effect by log-transforming the
KL-divergence, but up to this difference, the discrepancy here is the
same as the one used by \citet{Marttinen2015}. 
Also, because the discrepancy has been investigated before, we are able to construct \textit{a priori} plausible ranges for the parameters $\Theta = [0.01,0.15]\times[0.1,0.35]\times[4,10]$, and we use the uniform prior $p(\Btheta) = \Unif(\Theta)$. For the GP computations the \textit{hgt} parameter is scaled so that the parameters are approximately on the same scale. We run the simulation model in parallel with $1,000$ points generated from the prior. Most simulations require one to two hours on a single processor. We set the threshold to the $0.05$th quantile of the simulated discrepancy values, but the $0.01$th quantile led to similar conclusions. We model the discrepancy using the standard and input-dependent GP models and the same transformations as in the previous sections.

\begin{figure}
\centering
\includegraphics[width=\textwidth]{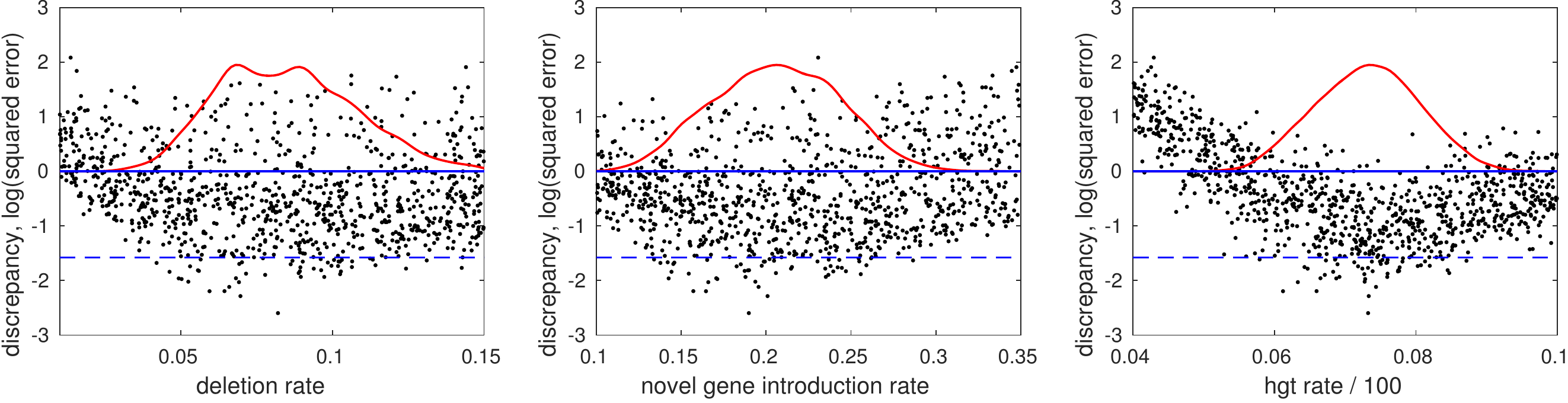}
\caption
{Marginal posterior densities for the three parameters of the genetics model. The discrepancy was log-transformed and the final model fitting was done by running the model $1000$ times and using the input-dependent GP model. The black dots are the model simulations (projected to each coordinate axis), the dashed blue line is the threshold and solid blue line describes the zero line. 
} \label{fig:gen_model_log_indep_gp_fit}
\end{figure}

The estimated posterior marginals are shown in Figure \ref{fig:gen_model_log_indep_gp_fit} and additional visualisation is included to the appendix \ref{sec:additional_genomics}. The largest \cmlpd{} score corresponds to the input-dependent GP model with the log transformation (\cmlpd{} $=-0.101$) but also the square-root transformation with input-dependent GP and the log transformation with standard GP yield visually similar approximations with utilities $-0.102$ and $-0.106$, respectively. On the other hand, the squared discrepancy in Equation \ref{eq:real_discrepancy} as such is difficult to model, resulting in overestimated posterior uncertainty (see Figure \ref{fig:intro_demo}). In general, the input-dependent GPs have higher utilities compared to the corresponding standard GPs for this simulation model. However, since we simulate only $1,000$ training data points, we expect the posterior variance to still be slightly overestimated, as seen in many toy examples. The approximated posterior agrees well with the earlier reported point estimate $\Btheta = (0.066, 0.18, 7.4)$. In addition, we see a strong positive correlation ($\rho=0.48$) between the \textit{del} and \textit{nov} parameters, which intuitively means that a high gene deletion rate can be compensated by a high rate of introducing novel genes into the population.

Finally, we derive posterior predictive distributions for two biologically interpretable quantities, i) the ratio between the number of all gene acquisitions vs.~gene deletions (computed by considering all acquisitions and deletions, caused either by HGT within the population or a novel acquisition/deletion), and ii) the ratio of gene introductions to the population from outside the population (as novel genes) vs. from within the population (through HGT). The posterior predictive distributions are obtained by re-weighting the original simulations with importance sampling. The $95\%$ credible interval for quantity i) is approximately $(1.17, 1.44)$ and for quantity ii) it is $(0.26, 0.52)$. Interestingly, we see that there are significantly more gene acquisitions than deletions, as with a high probability their ratio, the quantity i), is larger than one. Because in reality the genomes are not rapidly growing, this indicates some mechanism to counter the imbalance between acquisitions and deletions, for example selection against larger genomes in general, or alternatively that many new genes are individually selected against, see discussion by \citet{Marttinen2015}. On the other hand, the ratio of gene acquisitions from outside vs.~from within the population, the quantity ii), is approximately $0.4$, which corresponds to the biological expectation that the majority of horizontal gene transfer events happens between closely related bacterial strains, see e.g.~\citet{majewski2001sexual,fraser2007recombination}. To our knowledge, this has not been estimated before using simulation-based inference.

\section{Discussion} \label{sec:discussion}

We have thoroughly studied the use of GPs to enhance ABC inference, but, nevertheless, many choices could not be systematically investigated. We only considered the squared exponential covariance function, but expect the conclusions to hold also with other common options, as in \citet{Jabot2014}. We also used a zero mean function unlike \citet{Wilkinson2014,Gutmann2015}, who assumed that the discrepancy goes to infinity far from the minimum, and thus included quadratic terms to the mean function. Our choice allows for estimating posterior distributions of arbitrary shapes, at the cost of potentially overestimating the tails of the distributions. The results also depend on the GP hyperparameters; we used relatively uninformative priors for them and estimated them by maximising the marginal likelihood (for more details, see the appendix \ref{subsec:gp_model_details}). Integrating over the hyperparameters might improve the accuracy and stability, as in \citet{Snoek2012}, and alleviate numerical problems, which we occasionally encountered especially with the input-dependent GP model. Difficult cases included heavy-tailed, bimodal, or skewed discrepancy distributions, and cases where the discrepancy was approximately constant in some region but grew rapidly elsewhere.

We further assumed the summary statistics and the discrepancy function given, but in practice they must be designed carefully. We also considered a fixed set of transformations of the discrepancy, but other choices, such as the warped GP regression \citep{Snelson2004}, could be used to derive additional transformations. Overall, the error caused by a poorly designed discrepancy may be larger than the approximation error caused by an unsuitable GP model. Nevertheless, we find it important to understand and try to minimise the approximation error introduced in the modelling phase. In order to focus on the GP modelling aspect, we further assumed that the region with non-negligible posterior probability was known approximately in advance. In practice this could be estimated by Bayesian optimisation with the standard GP model. 
An interesting future direction is to formally integrate adaptive model selection with acquisition of novel evaluation locations.

While our study is the first to compare different models for the discrepancies, other studies on modelling in ABC have been conducted before. For example, \citet{Blum2010b} modelled individual summary statistics for regression adjustment method in ABC and allowed heteroscedastic noise. \citet{Blum2010} used different transformations of the summary statistics and investigated the selection of the corresponding regression adjustment method using cross-validation based criteria.
The normality assumptions of the synthetic likelihood method \citep{Wood2010} were examined by \citet{Price2016}, and, similarly to us, inferences were found relatively robust to deviations from normality, except when the summaries had heavy tails or were bimodal. \citet{Jabot2014} compared different emulation methods for ABC, namely local regressions and GPs. However, unlike in this work, the authors modelled the summaries separately, as was done also by \citet{Meeds2014}.

We applied the techniques to a previously published population genetic model for horizontal gene transfer in bacteria \citep{Marttinen2015}. In this realistic example, the input-dependent GP model with log-transformed discrepancies had the highest model selection utility, and was thus selected for presenting the results. This enabled us to derive the full posterior distribution for the parameters of the model. We estimated the number of gene acquisitions to be significantly higher than the number of gene deletions, suggesting some form of selection to prevent genomes from growing rapidly, to counterbalance this observation. We also estimated for the first time with simulation-based inference the ratio of gene transfers within the population considered, and those from external origins, and the results supported the empirical expectation that the majority of gene transfers happens between closely related strains. We note that multiple different models for bacterial evolution have been published, which differ in their purpose and assumptions \citep{fraser2007recombination,doroghazi2011model,cohan2007systematics,shapiro2012population,ansari2014inference,niehus2015migration}. The methods considered here establish a sound basis for estimating parameters in these models and their possible future generalisations.

\section{Conclusions} \label{sec:concl}

We considered the challenging task of ABC inference with a small number of model evaluations, and investigated the use of GPs to model the simulated discrepancies to fully use the scarce information available. Overall, we found this had a great potential to improve the accuracy of the posterior when the number of evaluations was limited. As anticipated by \citet{Gutmann2015}, we observed that the discrepancy distribution may in realistic situations deviate from standard GP assumptions, for example, the variance may be heteroscedastic or the distribution skewed or multimodal. For this reason, we studied various GP formulations for modelling the discrepancy, or the probability of the discrepancy being below the ABC threshold. 
We also investigated how transformations of the discrepancy affect the modelling accuracy.
The main finding is that no single modelling approach works best and, consequently, care is needed. 
Some general guidelines can be nevertheless be drawn:
\begin{itemize}
\item The input-dependent GP typically improves the results over the standard GP if the variance of the discrepancy is not constant across the parameter space. 
\item Square-root transformation produced the overall best approximations but also the log-transformation was often useful. 
However, squared discrepancy  
should be avoided due to its likely non-Gaussian distribution, and the dependence of the variance on the parameter, making it difficult to model with a GP.
\item Occasionally none of the GP models may fit the data well, leading to poor posterior approximations. In these cases the classifier GP, the smoothed ABC rejection sampler, or some more general GP formulation not included here may be useful.
\item Model selection tools can be used to select a GP model for ABC inference in a principled way, and their accuracy improves along with the number of model simulations available.
\end{itemize}

\section*{Acknowledgement}

This work was funded by the Academy of Finland (grants no. 286607 and 294015 to PM). We acknowledge the computational resources provided by the Aalto Science-IT project.

\bibliography{gp_abc_aoas_final.bib}
\bibliographystyle{plainnat}


\appendix \label{app:appendix}


\section{Additional details and results} \label{sec:supp}

\subsection{Further details on GP models} \label{subsec:gp_model_details}

We briefly describe the prior densities used for GP hyperparameters $\Bphi$. 
Generally speaking, we used rather noninformative priors. 
Specifically, for the standard GP model, we used the zero mean function i.e. $m(\Btheta) = 0$. For the log-transformation, however, we used a small negative mean function because a zero mean function in log-domain would correspond to $m(\Btheta) = \exp(0) = 1$ mean function in the original domain. We used t-distribution prior with location $0$, scale half of the range of the parameter space, and degrees of freedom (df) $4$ for each of the lengthscale parameters $l_i, i=1,\ldots,p$. In general, it can be difficult to know the scale and variation of the discrepancy across the parameter space. Thus we took a pragmatic approach and set a t-distribution prior with location=$0$, scale the standard deviation of the simulated discrepancies (computed using only trimmed values) and df=$4$ for $\sigma_f$. We used an improper uniform prior with support $\reals_{+}$ for $\sigma^2$. 

Compared to the standard GP model, the priors for the input-dependent GP model required more careful design to ensure robust computations and we used slightly more informative priors. Zero mean function was used i.e. $m(\Btheta) = 0$ (with the exception of log-transformation) as with standard GP. As mentioned in the main text, zero mean function was used also for $m_n(\Btheta)$. The lengthscale parameters are expected to have rather large values because we expect both the discrepancy and its variance to behave smoothly in all of our test cases. We also assume a priori that the variance of the discrepancy does not vary significantly over the parameter space. Consequently, we used t-distribution priors with the following (hyper)parameters: location: range divided by $3$, scale: range divided by $3$, df=$10$ for $l_{f_i}$; location: range divided by $2$, scale: range divided by $9$, df=$10$ for $l_{g_i}$; location: $0$, scale: standard deviation of the simulated discrepancies, df=$10$ for $\sigma_{f}$; and location=$0$, scale=$1$, df=$10$ for $\sigma_{g}$. 

For the binary GP classification, we set t-distribution prior with location: $0$, scale: the range of parameter space divided by $5$ and df=$4$ for each lengthscale $l_i$ and t-distribution prior with location=$0$, scale=$20$ and df=$4$ for the magnitude $\sigma_f$. 
Also, as discussed in the main text, the MAP estimate for the hyperparameters was used for all these GP models.

\subsection{GP model selection for ABC} \label{subsec:gp_model_sel}

\begin{figure}[H]
\centering
\includegraphics[width=.9\textwidth]{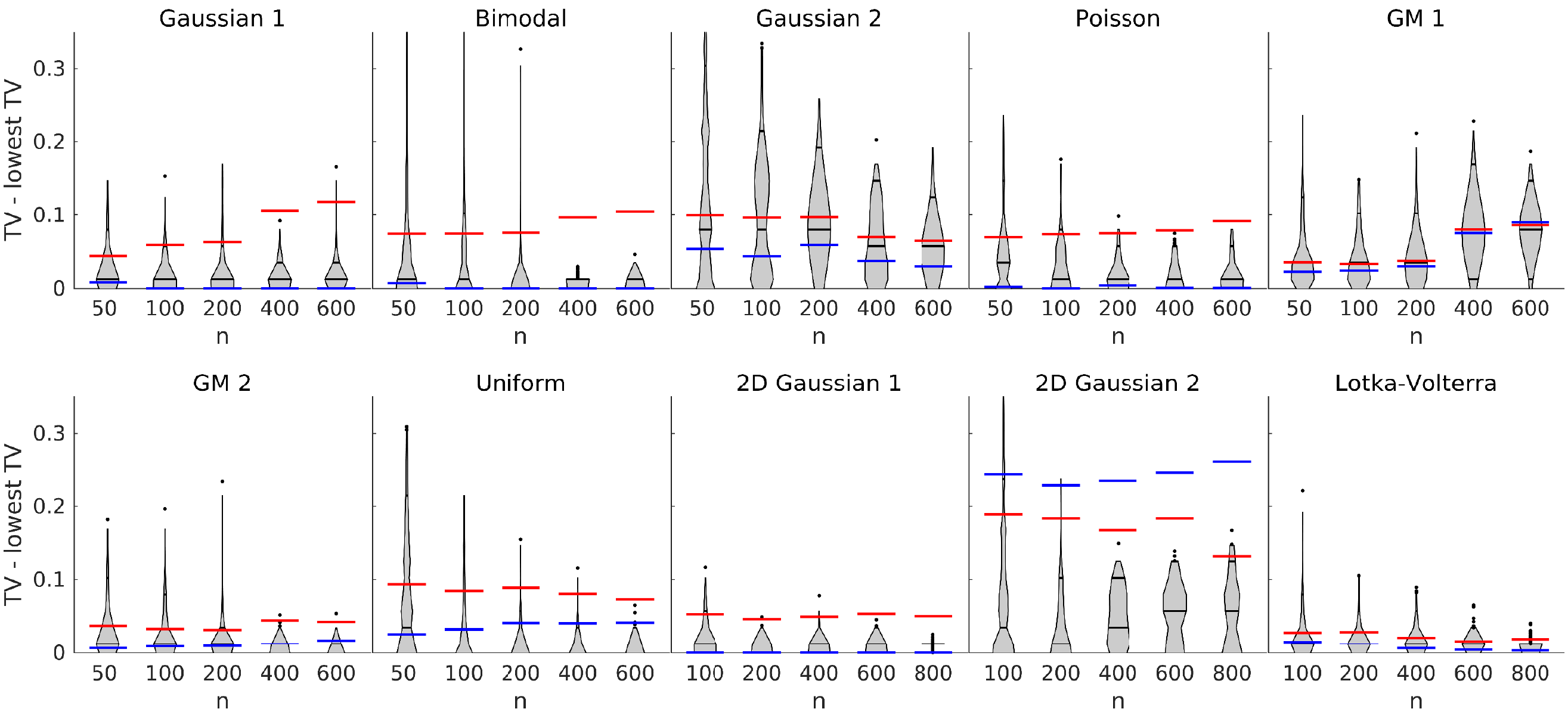}
\caption
{Results of the GP model selection using the \mlpd{}. 
See the caption of the Figure 6 in the main text for details. 
} \label{fig:mod_sel_gp_mlpd}
\end{figure}

We present the experimental results for the \mlpd{} in Figure \ref{fig:mod_sel_gp_mlpd}. As discussed in the main text, overall the results look similar to those of the \cmlpd{}, but with some test problems suboptimal GP models are systematically proposed.

\begin{figure}[H]
\centering
\includegraphics[width=.9\textwidth]{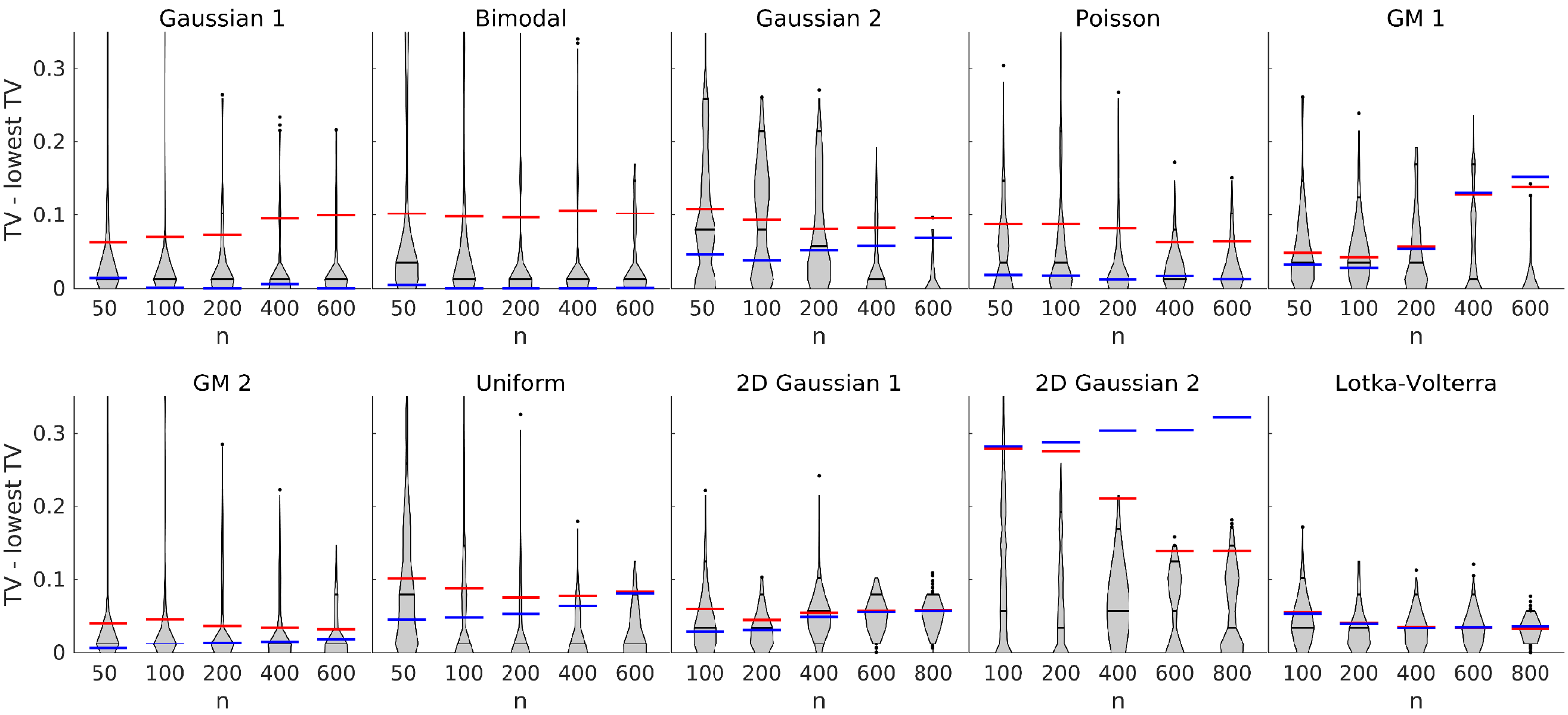}
\caption
{Results of the GP model selection using the \cmlpd{}. The experiments are the same as in Figure \ref{fig:mod_sel_gp_mlpd} expect that comparisons are made to the true posterior. 
} \label{fig:mod_sel_gp_logscore_truepost}
\end{figure}

\begin{figure}[H]
\centering
\includegraphics[width=.9\textwidth]{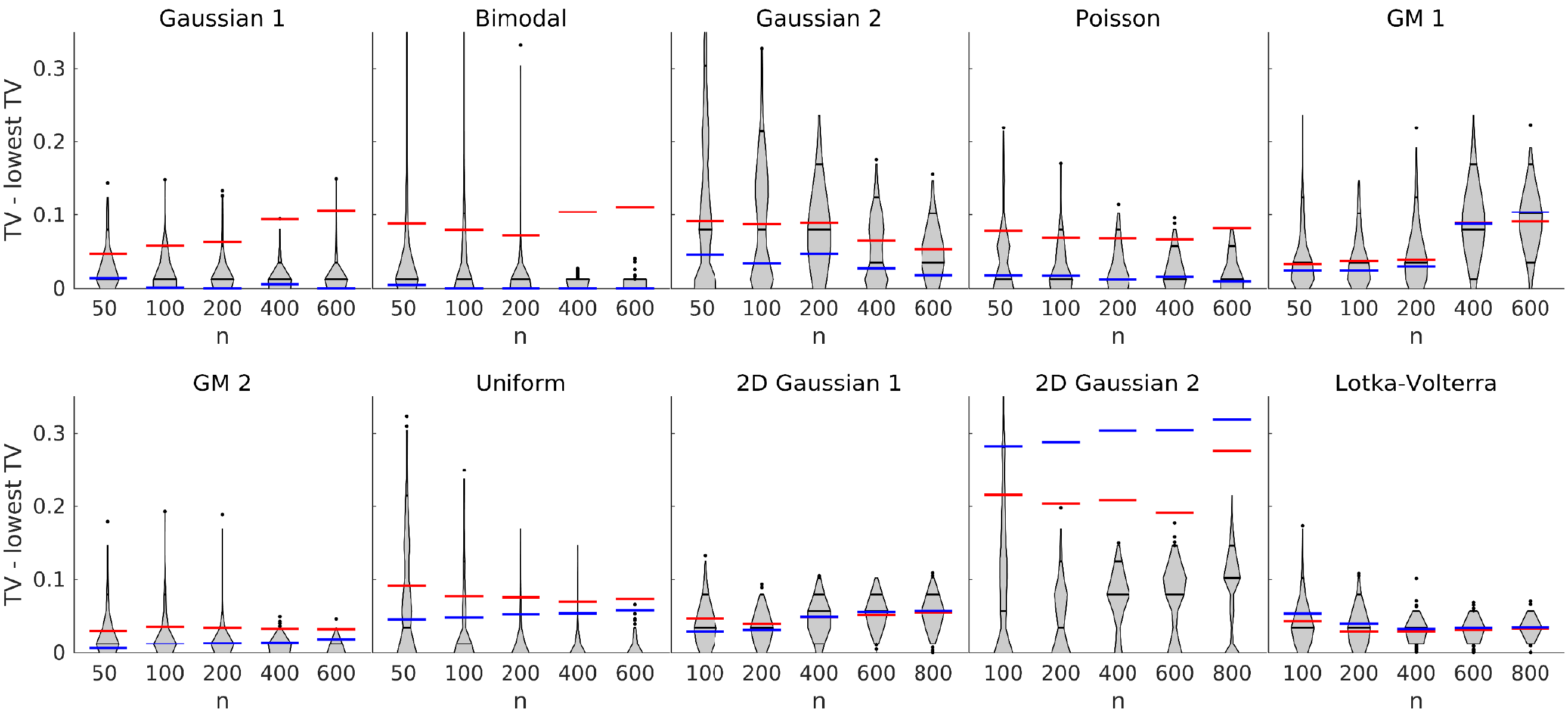}
\caption
{Results of the GP model selection using the \mlpd{}. The experiments are the same as in Figure \ref{fig:mod_sel_gp_mlpd} expect that comparisons are made to the true posterior. 
} \label{fig:mod_sel_gp_mlpd_truepost}
\end{figure}

Figures \ref{fig:mod_sel_gp_logscore_truepost} and \ref{fig:mod_sel_gp_mlpd_truepost} show the GP model selection results when the baseline is the true posterior instead of the ABC posterior that was used in the main text and in Figure \ref{fig:mod_sel_gp_mlpd}. 
Overall the results are similar but we observe some inconsistencies in the results of both utilities in the case of ``2D Gaussian 1'' and ``Lotka-Volterra'' test problems. This happens because the log-transformation produces too narrow posterior estimates compared to the corresponding true ABC posterior and thus poor approximations to it. However, the tendency to produce too narrow posterior estimates compensates the approximation error caused by the nonzero threshold and thus the log-transform produces the best estimates compared to the true posterior.

\subsection{Posterior estimates compared to the true posterior} \label{subsec:post_est_truepost}

Tables \ref{table:all_results1d_truepost} and \ref{table:all_results2d_truepost} contain the results of the experiments in the main text, but the baseline is the true posterior instead of the ABC posterior. 
Using the true posterior as a baseline does not change the main conclusions although posterior estimates are worse, as expected, because the nonzero threshold also affects the estimates. However, the additional error caused by the threshold is mostly minor in 1d test problems but in 2d test problems (especially with Lotka-Volterra model in which the discrepancy was a sum of squared errors of individual data points), the posterior estimates are clearly worse.

\begin{table}[H]
\ra{1.1}
\setlength{\tabcolsep}{4.0pt}
\scriptsize 
\begin{center}
\begin{tabular}{@{}llllllllllllllll@{}} 
\toprule
& \multicolumn{3}{c}{n=50} & \multicolumn{3}{c}{n=100} & \multicolumn{3}{c}{n=200} & \multicolumn{3}{c}{n=400} & \multicolumn{3}{c}{n=600} \\ 
\cmidrule{2-4} \cmidrule{5-7} \cmidrule{8-10} \cmidrule{11-13} \cmidrule{14-16} 
& se & log & sqrt & se & log & sqrt & se & log & sqrt & se & log & sqrt & se & log & sqrt \\ \midrule
Gaussian 1:\\
GP & 0.19 & 0.09 & \textbf{0.08} & 0.22 & 0.09 & \textbf{0.06} & 0.22 & 0.09 & \textbf{0.04} & 0.22 & 0.15 & \textbf{0.04} & 0.22 & 0.16 & \textbf{0.03} \\
GP in.dep. & 0.19 & 0.13 & 0.09 & 0.18 & 0.13 & \textbf{0.06} & 0.19 & 0.11 & 0.05 & 0.20 & 0.12 & \textbf{0.04} & 0.23 & 0.12 & 0.04 \\
\classgp{} & 0.41 & 0.41 & 0.41 & 0.34 & 0.34 & 0.34 & 0.19 & 0.19 & 0.19 & 0.11 & 0.11 & 0.11 & 0.09 & 0.09 & 0.09 \\
\abcrej{} & 0.32 & 0.32 & 0.32 & 0.27 & 0.27 & 0.27 & 0.19 & 0.19 & 0.19 & 0.12 & 0.12 & 0.12 & 0.12 & 0.12 & 0.12 \\ \hline
Bimodal:\\
GP & 0.22 & 0.48 & \textbf{0.16} & 0.21 & 0.26 & \textbf{0.12} & 0.22 & 0.17 & \textbf{0.10} & 0.22 & 0.15 & \textbf{0.07} & 0.22 & 0.16 & \textbf{0.07} \\
GP in.dep. & 0.19 & 0.60 & 0.18 & 0.17 & 0.40 & 0.14 & 0.21 & 0.27 & 0.11 & 0.21 & 0.24 & 0.08 & 0.21 & 0.22 & \textbf{0.07} \\
\classgp{} & 0.41 & 0.41 & 0.41 & 0.35 & 0.35 & 0.35 & 0.25 & 0.25 & 0.25 & 0.18 & 0.18 & 0.18 & 0.14 & 0.14 & 0.14 \\
\abcrej{} & 0.46 & 0.46 & 0.46 & 0.40 & 0.40 & 0.40 & 0.26 & 0.26 & 0.26 & 0.21 & 0.21 & 0.21 & 0.17 & 0.17 & 0.17 \\ \hline
Gaussian 2:\\
GP & 0.31 & 0.37 & \textbf{0.26} & 0.31 & \textbf{0.24} & 0.25 & 0.32 & \textbf{0.21} & 0.24 & 0.32 & 0.22 & 0.23 & 0.31 & 0.23 & 0.22 \\
GP in.dep. & 0.48 & 0.36 & 0.29 & 0.45 & 0.34 & 0.28 & 0.43 & 0.30 & 0.27 & 0.41 & 0.26 & 0.25 & 0.40 & 0.24 & 0.23 \\
\classgp{} & 0.35 & 0.35 & 0.35 & 0.34 & 0.34 & 0.34 & 0.25 & 0.25 & 0.25 & 0.18 & 0.18 & 0.18 & \textbf{0.15} & \textbf{0.15} & \textbf{0.15} \\
\abcrej{} & 0.28 & 0.28 & 0.28 & 0.29 & 0.29 & 0.29 & \textbf{0.21} & \textbf{0.21} & \textbf{0.21} & \textbf{0.17} & \textbf{0.17} & \textbf{0.17} & \textbf{0.15} & \textbf{0.15} & \textbf{0.15} \\ \hline
GM 1:\\
GP & 0.35 & \textbf{0.32} & 0.33 & 0.33 & \textbf{0.30} & 0.32 & 0.32 & 0.30 & 0.32 & 0.32 & 0.23 & 0.31 & 0.31 & 0.20 & 0.31 \\
GP in.dep. & 0.44 & 0.36 & 0.35 & 0.42 & 0.38 & 0.33 & 0.41 & 0.35 & 0.32 & 0.40 & 0.25 & 0.31 & 0.39 & 0.23 & 0.30 \\
\classgp{} & 0.36 & 0.36 & 0.36 & 0.37 & 0.37 & 0.37 & 0.29 & 0.29 & 0.29 & \textbf{0.19} & \textbf{0.19} & \textbf{0.19} & \textbf{0.15} & \textbf{0.15} & \textbf{0.15} \\
\abcrej{} & 0.37 & 0.37 & 0.37 & 0.34 & 0.34 & 0.34 & \textbf{0.25} & \textbf{0.25} & \textbf{0.25} & 0.20 & 0.20 & 0.20 & 0.16 & 0.16 & 0.16 \\ \hline
GM 2:\\
GP & 0.21 & 0.16 & \textbf{0.15} & 0.20 & \textbf{0.13} & 0.14 & 0.21 & \textbf{0.13} & \textbf{0.13} & 0.22 & \textbf{0.12} & 0.13 & 0.22 & \textbf{0.12} & 0.13 \\
GP in.dep. & 0.29 & 0.20 & \textbf{0.15} & 0.26 & 0.18 & 0.14 & 0.23 & 0.17 & \textbf{0.13} & 0.24 & 0.17 & \textbf{0.12} & 0.26 & 0.16 & \textbf{0.12} \\
\classgp{} & 0.38 & 0.38 & 0.38 & 0.33 & 0.33 & 0.33 & 0.21 & 0.21 & 0.21 & 0.16 & 0.16 & 0.16 & 0.15 & 0.15 & 0.15 \\
\abcrej{} & 0.34 & 0.34 & 0.34 & 0.30 & 0.30 & 0.30 & 0.22 & 0.22 & 0.22 & 0.18 & 0.18 & 0.18 & 0.16 & 0.16 & 0.16 \\ \hline
Uniform\\
GP & 0.28 & 0.19 & \textbf{0.17} & 0.29 & 0.20 & 0.16 & 0.29 & 0.19 & 0.17 & 0.29 & 0.20 & 0.17 & 0.29 & 0.20 & 0.17 \\
GP in.dep. & 0.27 & 0.20 & \textbf{0.17} & 0.24 & 0.20 & \textbf{0.13} & 0.21 & 0.20 & \textbf{0.12} & 0.19 & 0.20 & \textbf{0.12} & 0.17 & 0.20 & 0.11 \\
\classgp{} & 0.44 & 0.44 & 0.44 & 0.35 & 0.35 & 0.35 & 0.23 & 0.23 & 0.23 & 0.14 & 0.14 & 0.14 & \textbf{0.10} & \textbf{0.10} & \textbf{0.10} \\
\abcrej{} & 0.35 & 0.35 & 0.35 & 0.33 & 0.33 & 0.33 & 0.25 & 0.25 & 0.25 & 0.21 & 0.21 & 0.21 & 0.19 & 0.19 & 0.19 \\ \hline
Poisson:\\
GP & 0.21 & 0.13 & \textbf{0.09} & 0.20 & 0.09 & \textbf{0.08} & 0.20 & 0.08 & \textbf{0.07} & 0.22 & 0.09 & \textbf{0.07} & 0.22 & 0.10 & \textbf{0.07} \\
GP in.dep. & 0.20 & 0.22 & 0.12 & 0.22 & 0.18 & 0.09 & 0.25 & 0.16 & \textbf{0.07} & 0.24 & 0.14 & \textbf{0.07} & 0.25 & 0.14 & 0.08 \\
\classgp{} & 0.34 & 0.34 & 0.34 & 0.29 & 0.29 & 0.29 & 0.15 & 0.15 & 0.15 & 0.09 & 0.09 & 0.09 & 0.09 & 0.09 & 0.09 \\
\abcrej{} & 0.27 & 0.27 & 0.27 & 0.24 & 0.24 & 0.24 & 0.18 & 0.18 & 0.18 & 0.12 & 0.12 & 0.12 & 0.11 & 0.11 & 0.11 \\
\bottomrule
\end{tabular}
\caption{Results for the 1D toy examples. The numeric values describe the TV distance between the estimated and the corresponding true posterior density. 
} \label{table:all_results1d_truepost}
\end{center}
\end{table}

\begin{table}[H]
\ra{1.1}
\setlength{\tabcolsep}{4.0pt}
\scriptsize 
\begin{center}
\begin{tabular}{@{}llllllllllllllll@{}} 
\toprule
& \multicolumn{3}{c}{n=100} & \multicolumn{3}{c}{n=200} & \multicolumn{3}{c}{n=400} & \multicolumn{3}{c}{n=600} & \multicolumn{3}{c}{n=800} \\ 
\cmidrule{2-4} \cmidrule{5-7} \cmidrule{8-10} \cmidrule{11-13} \cmidrule{14-16}
& se & log & sqrt & se & log & sqrt & se & log & sqrt & se & log & sqrt & se & log & sqrt \\ \midrule
2D Gaussian 1:\\
GP & 0.35 & 0.19 & 0.19 & 0.34 & \textbf{0.15} & 0.18 & 0.33 & \textbf{0.11} & 0.16 & 0.33 & \textbf{0.10} & 0.16 & 0.33 & \textbf{0.10} & 0.15 \\
GP in.dep. & 0.27 & 0.20 & \textbf{0.17} & 0.26 & 0.17 & 0.16 & 0.26 & 0.15 & 0.15 & 0.28 & 0.14 & 0.14 & 0.28 & 0.14 & 0.14 \\ 
\classgp{} & 0.69 & 0.69 & 0.69 & 0.70 & 0.70 & 0.70 & 0.29 & 0.29 & 0.29 & 0.21 & 0.21 & 0.21 & 0.19 & 0.19 & 0.19 \\
\abcrej{} & 0.43 & 0.43 & 0.43 & 0.35 & 0.35 & 0.35 & 0.30 & 0.30 & 0.30 & 0.27 & 0.27 & 0.27 & 0.25 & 0.25 & 0.25 \\ \hline
2D Gaussian 2:\\
GP & 0.62 & 0.44 & 0.56 & 0.62 & 0.39 & 0.54 & 0.61 & 0.35 & 0.52 & 0.61 & 0.34 & 0.52 & 0.61 & 0.33 & 0.52 \\
GP in.dep. & 0.72 & \textbf{0.30} & 0.34 & 0.72 & \textbf{0.25} & 0.33 & 0.56 & \textbf{0.22} & 0.32 & 0.56 & \textbf{0.21} & 0.32 & 0.56 & \textbf{0.20} & 0.31 \\
\classgp{} & 0.70 & 0.70 & 0.70 & 0.71 & 0.71 & 0.71 & 0.44 & 0.44 & 0.44 & 0.28 & 0.28 & 0.28 & 0.24 & 0.24 & 0.24 \\
\abcrej{} & 0.45 & 0.45 & 0.45 & 0.40 & 0.40 & 0.40 & 0.34 & 0.34 & 0.34 & 0.32 & 0.32 & 0.32 & 0.31 & 0.31 & 0.31 \\ \hline
Lotka-Volterra:\\
GP & 0.58 & 0.52 & 0.55 & 0.57 & 0.49 & 0.52 & 0.56 & 0.48 & 0.51 & 0.56 & \textbf{0.47} & 0.51 & 0.56 & \textbf{0.47} & 0.50 \\
GP in.dep. & 0.65 & \textbf{0.50} & 0.53 & 0.56 & \textbf{0.48} & 0.51 & 0.55 & \textbf{0.47} & 0.50 & 0.55 & \textbf{0.47} & 0.50 & 0.55 & \textbf{0.47} & 0.50 \\ 
\classgp{} & 0.79 & 0.79 & 0.79 & 0.79 & 0.79 & 0.79 & 0.79 & 0.79 & 0.79 & 0.54 & 0.54 & 0.54 & 0.51 & 0.51 & 0.51 \\
\abcrej{} & 0.65 & 0.65 & 0.65 & 0.59 & 0.59 & 0.59 & 0.55 & 0.55 & 0.55 & 0.56 & 0.56 & 0.56 & 0.54 & 0.54 & 0.54 \\
\bottomrule
\end{tabular}
\caption{Results for the 2D toy examples. The numeric values describe the TV distance between the estimated and the corresponding true posterior density. } \label{table:all_results2d_truepost}
\end{center}
\end{table}

\subsection{Selection of the threshold} \label{subsec:threshold}

The 0.05th quantile of the simulated discrepancies was chosen for the experiments shown in the main text and for the additional experiments in Sections \ref{subsec:gp_model_sel} and \ref{subsec:post_est_truepost}. However, we also repeated the experiments with the 0.01th quantile and the corresponding results are shown in Tables \ref{table:all_results1d_001q} and \ref{table:all_results2d_001q}. The ABC posterior is used as the baseline here. 
In general, the results are similar to those with the 0.05th quantile. However, as expected, the classification GP and ABC rejection sampler methods perform much worse due to the smaller number of discrepancy realisations below the threshold. 
The computations were also repeated with the 0.1th quantile and the results were generally similar to those using the 0.05th quantile, except that the error was larger when compared to the true posterior. Consequently, these results are not reported here.

\begin{table}[H] 
\ra{1.1}
\setlength{\tabcolsep}{4.0pt}
\scriptsize 
\begin{center}
\begin{tabular}{@{}llllllllllllllll@{}} 
\toprule
& \multicolumn{3}{c}{n=50} & \multicolumn{3}{c}{n=100} & \multicolumn{3}{c}{n=200} & \multicolumn{3}{c}{n=400} & \multicolumn{3}{c}{n=600} \\ 
\cmidrule{2-4} \cmidrule{5-7} \cmidrule{8-10} \cmidrule{11-13} \cmidrule{14-16} 
& se & log & sqrt & se & log & sqrt & se & log & sqrt & se & log & sqrt & se & log & sqrt \\ \midrule
Gaussian 1:\\
GP & 0.18 & 0.20 & \textbf{0.08} & 0.21 & 0.23 & \textbf{0.06} & 0.21 & 0.24 & \textbf{0.05} & 0.21 & 0.32 & \textbf{0.04} & 0.21 & 0.32 & \textbf{0.03} \\
GP in.dep. & 0.20 & 0.28 & 0.10 & 0.18 & 0.31 & 0.07 & 0.19 & 0.29 & 0.06 & 0.20 & 0.30 & 0.05 & 0.22 & 0.30 & 0.05 \\
\classgp{} & 0.51 & 0.51 & 0.51 & 0.56 & 0.56 & 0.56 & 0.56 & 0.56 & 0.56 & 0.52 & 0.52 & 0.52 & 0.39 & 0.39 & 0.39 \\
\abcrej{} & 0.60 & 0.60 & 0.60 & 0.60 & 0.60 & 0.60 & 0.37 & 0.37 & 0.37 & 0.26 & 0.26 & 0.26 & 0.20 & 0.20 & 0.20 \\ \hline
Bimodal:\\
GP & 0.21 & 0.53 & \textbf{0.16} & 0.21 & 0.45 & \textbf{0.13} & 0.21 & 0.31 & \textbf{0.11} & 0.21 & 0.31 & \textbf{0.09} & 0.21 & 0.33 & \textbf{0.08} \\
GP in.dep. & 0.18 & 0.68 & 0.20 & 0.17 & 0.60 & 0.15 & 0.21 & 0.49 & 0.12 & 0.21 & 0.46 & 0.10 & 0.21 & 0.43 & 0.09 \\
\classgp{} & 0.44 & 0.44 & 0.44 & 0.44 & 0.44 & 0.44 & 0.46 & 0.46 & 0.46 & 0.43 & 0.43 & 0.43 & 0.38 & 0.38 & 0.38 \\
\abcrej{} & 0.47 & 0.47 & 0.47 & 0.47 & 0.47 & 0.47 & 0.48 & 0.48 & 0.48 & 0.44 & 0.44 & 0.44 & 0.32 & 0.32 & 0.32 \\ \hline
Gaussian 2:\\
GP & 0.32 & 0.46 & \textbf{0.27} & 0.32 & 0.31 & \textbf{0.25} & 0.32 & 0.28 & \textbf{0.25} & 0.32 & 0.30 & \textbf{0.23} & 0.32 & 0.31 & \textbf{0.23} \\
GP in.dep. & 0.48 & 0.48 & 0.35 & 0.45 & 0.50 & 0.35 & 0.44 & 0.45 & 0.32 & 0.42 & 0.43 & 0.30 & 0.41 & 0.41 & 0.30 \\
\classgp{} & 0.45 & 0.45 & 0.45 & 0.43 & 0.43 & 0.43 & 0.47 & 0.47 & 0.47 & 0.47 & 0.47 & 0.47 & 0.44 & 0.44 & 0.44 \\
\abcrej{} & 0.48 & 0.48 & 0.48 & 0.48 & 0.48 & 0.48 & 0.36 & 0.36 & 0.36 & 0.29 & 0.29 & 0.29 & 0.27 & 0.27 & 0.27 \\ \hline
GM 1:\\
GP & 0.34 & \textbf{0.32} & \textbf{0.32} & 0.32 & \textbf{0.31} & \textbf{0.31} & \textbf{0.31} & 0.32 & \textbf{0.31} & 0.31 & 0.30 & 0.30 & 0.30 & \textbf{0.29} & \textbf{0.29} \\
GP in.dep. & 0.43 & 0.48 & 0.34 & 0.41 & 0.54 & 0.32 & 0.40 & 0.49 & \textbf{0.31} & 0.39 & 0.43 & \textbf{0.29} & 0.38 & 0.41 & \textbf{0.29} \\
\classgp{} & 0.42 & 0.42 & 0.42 & 0.43 & 0.43 & 0.43 & 0.43 & 0.43 & 0.43 & 0.42 & 0.42 & 0.42 & 0.38 & 0.38 & 0.38 \\
\abcrej{} & 0.43 & 0.43 & 0.43 & 0.43 & 0.43 & 0.43 & 0.39 & 0.39 & 0.39 & 0.34 & 0.34 & 0.34 & \textbf{0.29} & \textbf{0.29} & \textbf{0.29} \\ \hline
GM 2:\\
GP & 0.19 & 0.18 & \textbf{0.13} & 0.19 & 0.22 & \textbf{0.12} & 0.19 & 0.22 & \textbf{0.12} & 0.21 & 0.22 & 0.12 & 0.21 & 0.22 & 0.12 \\
GP in.dep. & 0.28 & 0.31 & 0.14 & 0.25 & 0.34 & 0.13 & 0.22 & 0.30 & \textbf{0.12} & 0.23 & 0.28 & \textbf{0.11} & 0.25 & 0.27 & \textbf{0.10} \\
\classgp{} & 0.43 & 0.43 & 0.43 & 0.46 & 0.46 & 0.46 & 0.47 & 0.47 & 0.47 & 0.47 & 0.47 & 0.47 & 0.33 & 0.33 & 0.33 \\
\abcrej{} & 0.49 & 0.49 & 0.49 & 0.49 & 0.49 & 0.49 & 0.36 & 0.36 & 0.36 & 0.28 & 0.28 & 0.28 & 0.25 & 0.25 & 0.25 \\ \hline
Uniform\\
GP & 0.27 & 0.35 & \textbf{0.16} & 0.28 & 0.38 & 0.16 & 0.28 & 0.37 & 0.17 & 0.28 & 0.38 & 0.17 & 0.28 & 0.38 & 0.17 \\
GP in.dep. & 0.27 & 0.31 & 0.17 & 0.23 & 0.33 & \textbf{0.13} & 0.21 & 0.35 & \textbf{0.13} & 0.19 & 0.39 & \textbf{0.12} & 0.18 & 0.38 & \textbf{0.12} \\
\classgp{} & 0.45 & 0.45 & 0.45 & 0.49 & 0.49 & 0.49 & 0.51 & 0.51 & 0.51 & 0.51 & 0.51 & 0.51 & 0.48 & 0.48 & 0.48 \\
\abcrej{} & 0.51 & 0.51 & 0.51 & 0.51 & 0.51 & 0.51 & 0.42 & 0.42 & 0.42 & 0.33 & 0.33 & 0.33 & 0.30 & 0.30 & 0.30 \\ \hline
Poisson:\\
GP & 0.20 & 0.21 & \textbf{0.10} & 0.19 & 0.25 & \textbf{0.08} & 0.20 & 0.25 & \textbf{0.07} & 0.21 & 0.29 & \textbf{0.06} & 0.21 & 0.31 & \textbf{0.07} \\
GP in.dep. & 0.20 & 0.39 & 0.14 & 0.21 & 0.40 & 0.12 & 0.26 & 0.38 & 0.09 & 0.25 & 0.37 & 0.10 & 0.26 & 0.38 & 0.10 \\
\classgp{} & 0.45 & 0.45 & 0.45 & 0.50 & 0.50 & 0.50 & 0.38 & 0.38 & 0.38 & 0.28 & 0.28 & 0.28 & 0.18 & 0.18 & 0.18 \\
\abcrej{} & 0.57 & 0.57 & 0.57 & 0.43 & 0.43 & 0.43 & 0.26 & 0.26 & 0.26 & 0.19 & 0.19 & 0.19 & 0.16 & 0.16 & 0.16 \\
\bottomrule
\end{tabular}
\caption{Results for the 1D toy examples when the threshold is 0.01th quantile. The numeric values describe the TV distance between the estimated and the corresponding ABC posterior density. 
} \label{table:all_results1d_001q}
\end{center}
\end{table}

\begin{table}[H]
\ra{1.1}
\setlength{\tabcolsep}{4.0pt}
\scriptsize 
\begin{center}
\begin{tabular}{@{}llllllllllllllll@{}} 
\toprule
& \multicolumn{3}{c}{n=100} & \multicolumn{3}{c}{n=200} & \multicolumn{3}{c}{n=400} & \multicolumn{3}{c}{n=600} & \multicolumn{3}{c}{n=800} \\ 
\cmidrule{2-4} \cmidrule{5-7} \cmidrule{8-10} \cmidrule{11-13} \cmidrule{14-16}
& se & log & sqrt & se & log & sqrt & se & log & sqrt & se & log & sqrt & se & log & sqrt \\ \midrule
2D Gaussian 1:\\
GP & 0.29 & 0.29 & \textbf{0.12} & 0.28 & 0.28 & \textbf{0.09} & 0.27 & 0.28 & \textbf{0.07} & 0.27 & 0.28 & \textbf{0.07} & 0.27 & 0.29 & \textbf{0.06} \\
GP in.dep. & 0.24 & 0.41 & 0.16 & 0.24 & 0.33 & 0.11 & 0.23 & 0.28 & 0.09 & 0.25 & 0.27 & 0.08 & 0.25 & 0.27 & 0.07 \\
\classgp{} & 0.65 & 0.65 & 0.65 & 0.67 & 0.67 & 0.67 & 0.67 & 0.67 & 0.67 & 0.67 & 0.67 & 0.67 & 0.67 & 0.67 & 0.67 \\
\abcrej{} & 0.68 & 0.68 & 0.68 & 0.68 & 0.68 & 0.68 & 0.37 & 0.37 & 0.37 & 0.34 & 0.34 & 0.34 & 0.31 & 0.31 & 0.31 \\ \hline
2D Gaussian 2:\\
GP & 0.59 & \textbf{0.38} & 0.52 & 0.59 & \textbf{0.28} & 0.49 & 0.58 & \textbf{0.25} & 0.47 & 0.57 & \textbf{0.25} & 0.46 & 0.57 & \textbf{0.25} & 0.46 \\
GP in.dep. & 0.69 & 0.50 & 0.41 & 0.67 & 0.39 & 0.37 & 0.59 & 0.35 & 0.36 & 0.59 & 0.31 & 0.37 & 0.59 & 0.29 & 0.35 \\
\classgp{} & 0.68 & 0.68 & 0.68 & 0.68 & 0.68 & 0.68 & 0.69 & 0.69 & 0.69 & 0.69 & 0.69 & 0.69 & 0.68 & 0.68 & 0.68 \\
\abcrej{} & 0.70 & 0.70 & 0.70 & 0.70 & 0.70 & 0.70 & 0.41 & 0.41 & 0.41 & 0.36 & 0.36 & 0.36 & 0.32 & 0.32 & 0.32 \\ \hline
Lotka-Volterra:\\
GP & 0.28 & \textbf{0.24} & 0.25 & 0.24 & \textbf{0.20} & \textbf{0.20} & 0.21 & \textbf{0.18} & \textbf{0.18} & 0.20 & 0.17 & \textbf{0.16} & 0.19 & 0.16 & \textbf{0.14} \\
GP in.dep. & 0.53 & 0.31 & 0.30 & 0.29 & 0.27 & 0.24 & 0.25 & 0.23 & 0.20 & 0.24 & 0.22 & 0.18 & 0.22 & 0.20 & 0.16 \\
\classgp{} & 0.59 & 0.59 & 0.59 & 0.60 & 0.60 & 0.60 & 0.59 & 0.59 & 0.59 & 0.59 & 0.59 & 0.59 & 0.59 & 0.59 & 0.59 \\
\abcrej{} & 0.60 & 0.60 & 0.60 & 0.60 & 0.60 & 0.60 & 0.35 & 0.35 & 0.35 & 0.31 & 0.31 & 0.31 & 0.29 & 0.29 & 0.29 \\
\bottomrule
\end{tabular}
\caption{Results for the 2D toy examples when the threshold is the 0.01th quantile. The numeric values describe the TV distance between the estimated and the corresponding ABC posterior density.} \label{table:all_results2d_001q}
\end{center}
\end{table}


\newpage
\section{Bivariate posterior marginals for the bacterial genomic model} \label{sec:additional_genomics}
${}$

\begin{figure}[H]
\centering
\includegraphics[width=.65\textwidth]{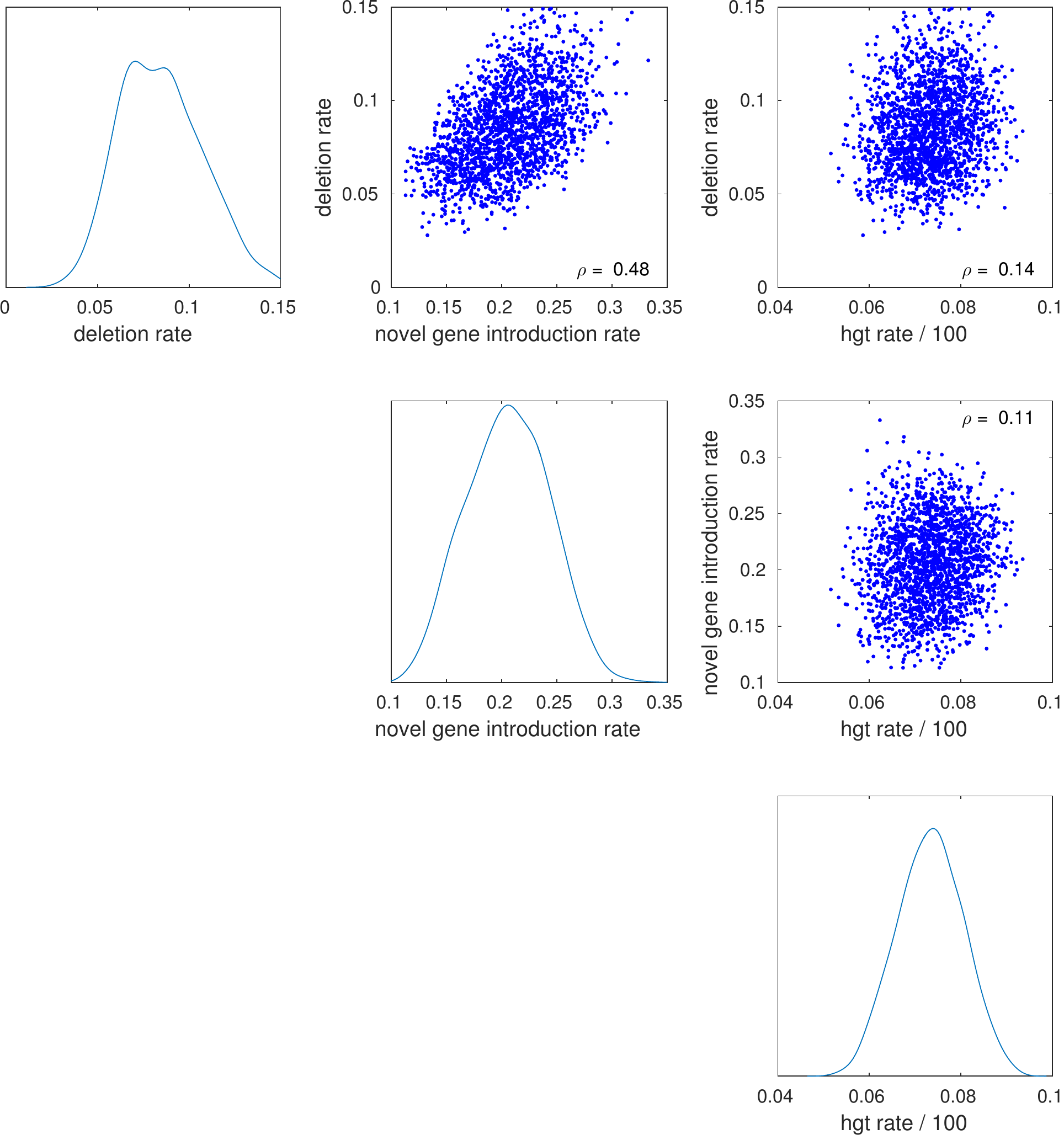}
\caption
{Estimated posterior distribution for the bacterial genomic model. The value of $\rho$ is the (Pearson) correlation coefficient between the corresponding parameters of the model. 
} \label{fig:gen3x3}
\end{figure}

The bivariate marginal posteriors for the bacterial genomic model described in Section 3.3 of the main text are shown in Figure \ref{fig:gen3x3}. 

\label{lastpage}

\end{document}